\documentclass[journal, letterpaper]{IEEEtran}

\usepackage{graphicx}
\usepackage{amsmath}    
\usepackage{textgreek}	% Greek to me, dawg
\usepackage{listings}
\usepackage{csvsimple}
\usepackage{longtable}
\usepackage{authblk}
\usepackage{url}
\usepackage{caption}
\usepackage{multirow}
\usepackage{threeparttable}
\usepackage{cite}

\begin{document}
\title{Natural Language Instructions for Intuitive Human Interaction with Robotic Assistants in Field Construction Work}

    \author{Somin Park$^1$\thanks{$^1$Dept. of Civil and Env. Engineering, University of Michigan \{somin, menassa, vkamat\}@umich.edu.}, Xi Wang$^2$\thanks{$^2$Dept. of Construction Science, Texas A\&M University, xiwang@tamu.edu.}, Carol C. Menassa$^1$, Vineet R. Kamat$^1$, and Joyce Y. Chai$^3$\thanks{$^3$Dept. of Elec. Engineering and Computer Science, University of Michigan, chaijy@umich.edu.}}
    \maketitle

\begin{abstract}
	The introduction of robots is widely considered to have significant potential of alleviating the issues of worker shortage and stagnant productivity that afflict the construction industry. However, it is challenging to use fully automated robots in complex and unstructured construction sites. Human-Robot Collaboration (HRC) has shown promise of combining human workers’ flexibility and robot assistants’ physical abilities to jointly address the uncertainties inherent in construction work. When introducing HRC in construction, it is critical to recognize the importance of teamwork and supervision in field construction and establish a natural and intuitive communication system for the human workers and robotic assistants. Natural language-based interaction can enable intuitive and familiar communication with robots for human workers who are non-experts in robot programming. However, limited research has been conducted on this topic in construction. This paper proposes a framework to allow human workers to interact with construction robots based on natural language instructions. The proposed method consists of three stages: Natural Language Understanding (NLU), Information Mapping (IM), and Robot Control (RC). Natural language instructions are input to a language model to predict a tag for each word in the NLU module. The IM module uses the result of the NLU module and building component information to generate the final instructional output essential for a robot to acknowledge and perform the construction task. A case study for drywall installation is conducted to evaluate the proposed approach. The results of the NLU and IM modules show high accuracy over 99\%, allowing a robot to perform tasks accurately for a given set of natural language instructions in the RC module. The obtained results highlight the potential of using natural language-based interaction to replicate the communication that occurs between human workers within the context of human-robot teams. 
\end{abstract}

\section{Introduction}
Robots have been adopted in the construction industry to support diverse field activities such as bricklaying  \cite{Wu1}, earthmoving \cite{azar1}, painting \cite{grassi1}, underground exploration \cite{moon1}, concrete placement \cite{wi}, tunnel inspection \cite{menendez}, curtain wall assembly \cite{taghavi}, and wall-cleaning \cite{joo}. Robotics is considered an effective means to address issues of labor shortages and stagnant growth of productivity in construction \cite{delgado, lundeen2019, cai}. However, it is challenging for robots to work on construction sites due to evolving and unstructured work environments \cite{feng, lundeen2017}, differing conditions from project to project \cite{pan}, and the prevalence of quasi-repetitive work tasks \cite{liang}. This is in contrast to automated manufacturing facilities that have structured environments \cite{feng}. It is expected that the construction robots will encounter situations different than what are stipulated in the design documents and will have to work with the human collaborator to resolve such unexpected conditions. Collaboration between humans and robots has the potential to address several such challenges inherent in the performance of construction tasks in the field. The advantage of collaborative robots lies in the opportunity to combine human intelligence and flexibility with robot strength, precision, and repeatability \cite{michalos, sherwani}. Collaboration can increase productivity and quality of the construction tasks and human safety \cite{su, pini}. It can also reduce physical exertion for humans since repetitive tasks will be carried out by robots. Therefore, in Human-Robot Collaboration (HRC), skills of human operators and robots can complement each other to complete designated tasks. 

In construction, communication between teammates is essential since construction work crews have many degrees of freedom in organizing and coordinating the work, and dynamic and unpredictable environments create high likelihood of errors \cite{cupido}. Similarly, when collaborative robots assist human workers, interaction between humans and robots is critical in the construction field \cite{delgado}. In human-robot construction teams, most of the robots are currently in the lower level of robot autonomy where human workers determine task plans and robots execute them \cite{liang}. To deliver plans generated by human workers to robots, human operators need proper interfaces \cite{wang2021kamat}. However, designing intuitive user interfaces is one of the key challenges of HRC since interaction with robots usually requires specialized knowledge in humans \cite{villani}. Intuitive and natural interaction enables human operators to easily interact with robots and take full advantage of human skills, resulting in enhanced productivity \cite{maurtua, villani}. In addition, during the natural interaction, shallower learning curves can be expected with future novice operators and low levels of fatigue can be maintained. Therefore, it is important to establish a natural and intuitive communication approach to achieve successful HRC in the construction industry.

In a natural interaction-based workflow, humans can interact efficiently with robots as they communicate with other human workers, and the robots can be endowed with the capability to capture and accurately process human requests and then carry out a series of tasks. Several recent studies have investigated natural HRC in the construction industry using various communication channels such as gesture \cite{tanzini}, Virtual Reality (VR) \cite{adami}, brainwaves \cite{liu2021}, and speech \cite{follini}. Among them, speech interaction has been considered as the most natural and intuitive way of communication in the human-robot interaction field \cite{karpaga,tsarou, peng, mukherjee}. HRC using voice commands helps human operators focus on tasks since hands-free interaction is possible and the operators’ mobility is not restricted \cite{villani}. In industrial settings including construction sites, noisy environments can affect speech recognition. However, with recent advances in noise-robust speech recognition, it is expected that using voice input commands is feasible in noisy environments \cite{qian2016, fukumori}.

Natural language-based interaction, in which speech input is used, has attracted increasing attention with its advantages in the field of robotics \cite{hatori, ye}. Using natural language instructions allows human operators to deliver their requests accurately and efficiently \cite{liuandzhang2017}. Users’ intents about action, tools, workpieces, and location for HRC can be accurately expressed through natural language without information loss in ways distinct from other simplified requests \cite{liu2016, paul2016}. In addition, users do not need to design informative expressions when communicating through existing languages, making the interaction efficient. Given these advantages, language instructions have been used to make robots perform pick-and-place operations, one of the most common tasks of industrial robots \cite{paul2016, bisk2016, hatori, magass2019}.

For pick-and-place operations to install or assemble construction workpieces, the workpieces can be described by their IDs or characteristics such as dimension, position, or material available from project information (e.g., Building Information Model). Most of the existing methods for analyzing language instructions for a pick-and-place operation have extracted information about its final location and workpieces described by its color, name, or spatial relationships for household tasks \cite{bisk2016, hatori, magass2019}. For construction tasks where the same materials are used repeatedly, color or name may not be reliable features for indicating objects without ambiguity. Therefore, it is essential to use precise workpiece descriptions in language instructions for construction tasks, such as quantitative dimensions, IDs, positions, and previous working records. In addition, in robotic planning in construction, orientation of a target object is one of the essential information for automated placement planning of building components \cite{chong2022}. 

\subsection{Research Objectives}
The investigation of HRC in construction field, particularly in relation to natural language usage, requires further exploration. This study aims to bridge this gap by proposing a framework for natural interaction with construction robots through the use of natural language instructions and building component information. Specifically, the focus is on analyzing natural language instructions for pick-and-place construction operations within a low-level HRC context. To address the scarcity of resources in terms of natural language instruction datasets in the construction industry, a fine-grained annotation is created. This annotation enables the identification of unique workpiece characteristics and allows for detailed analysis. By incorporating this detailed annotation, it is anticipated that the quality and depth of the labeled data can be enhanced while it introduces a higher degree of complexity to language understanding.

To validate the effectiveness of the proposed approach, this study involves the training and comparison of two existing language models using new datasets. The results obtained from these models are then applied to the building component information available in construction projects. Moreover, a set of experiments on drywall installation is conducted as a case study to demonstrate and evaluate the proposed approach.

\section{Literature Review}
Through the review of existing works, the need for this study and research gaps are identified. The first section establishes the need for analyzing natural language instructions for HRC of the construction domain. The second section examines the characteristics of data and approach used in other domains in relation to natural language understanding. The third section investigates studies that performed information extraction in the construction industry.

\subsection{Interaction between human workers and robots in the construction industry}
Advanced interaction methods for HRC enable human workers to collaborate with robots easily and naturally. In construction, research using gestures, VR, brain signals, and speech has been proposed for interaction with robots. Gesture-based interaction using operators’ body movements can enhance the intuitiveness of communication \cite{albeaino2022} and be used in noisy environments such as construction sites \cite{wang2021}. Wang and Zhu \cite{wang2021} proposed a vision-based framework for interpreting nine hand gestures to control construction machines. Sensor-based wearable glove systems were proposed to recognize hand gestures for driving hydraulic machines \cite{tanzini} and loaders \cite{tiesen2020}. However, when using hand gestures, the operators’ hands are not free, and they have to keep pointing to the endpoint, which may lead to fatigue \cite{tolgy2017}.

VR interfaces have been used in the construction industry for visual simulation, building reconnaissance, worker training, safety management system, labor management and other applications (e.g., \cite{kamat2002, kamat2003, dong2013, ahmed2017}). It can also provide an opportunity for users to control robots without safety risks \cite{perez2019}. Regarding interaction with robots, Zhou et al. \cite{zhou2020} and Wang et al. \cite{wang2021kamat} tested VR as an intuitive user interface exploring the virtual scene for pipe operation and drywall installation, respectively. Both studies sent commands to robots by handheld controllers, which determined desired poses and actions of robots. In addition to the purpose of operating robots, Adami et al. \cite{adami} investigated the impacts of VR-based training for remote-operating construction robots. In the interaction with a demolition robot, operators used the robot’s controller consisting of buttons and joysticks based on digital codes. However, head mounted devices as visual displays may be uncomfortable for operators due to onset of eye strain and hand-held devices may limit the operators in their actions \cite{behzadan2011, dalle2021}. In addition, the connection between the headset and the controllers can be interrupted, and the working space is limited due to cables attached to the computer \cite{diana2021}.

Recently, brain-control methods have been proposed for HRC in construction, translating the signals into a set of commands for robots. To control robots, users can attempt to convey their intention in a direct and natural way by manipulating their brain activities \cite{ji2021}. In construction, Liu et al. \cite{liu2021} and Liu et al. \cite{liu2021_b} proposed systems for brain-computer interfaces to allow human workers to implement hands-free control of robots. Users’ brainwaves were captured from an electroencephalogram (EEG) and interpreted into three directional commands (left, right, and stop) \cite{liu2021}. In the other study \cite{liu2021_b}, brainwaves were classified into three levels of cognitive load (low, medium, and high), and the results were used for robotic adjustment. This communication using brain signals enables physiologically-based HRC by evaluating workers’ mental states \cite{liu2021_b}. However, systems using brain signals have to overcome challenges of time consumption for user training, non-stationarity of signals affected by the mental status of users, and user discomfort by moist sensors using a gel \cite{aljalal2020}. It is also challenging for users to deliver high-dimensional commands to collaborative robots because of the limited number of classifiable mental states \cite{ji2021}.

Speech is the most natural way of communication in humans, even if the objects of their communication are not other humans but machines or computers \cite{karpaga, peng}. It is a flexible medium for construction workers to communicate with robots, which can be leveraged for hands-free and eyes-free interaction with low-level training \cite{abioye2021}. Even if noisy construction sites could generate many errors in verbal communication, it has the potential to be used in noisy environments with recent advances in noise-robust speech recognition \cite{qian2016, fukumori}. Natural language is important in human-human interaction during teamwork since it helps seamless communication. Enabling robots to understand natural language commands also facilitates flexible communication in human-robot teams \cite{beetz2017}. Untrained users can effectively control robots in a natural and intuitive way using natural language. Despite the advantages of the speech channel and natural language in interaction, there are few studies examining natural language instructions for human-robot collaboration in construction. Follini et al. \cite{follini} proposed a robotic gripper system integrated with voice identification/authentication for automated scaffolding assembly, but it was based on a very limited number of simple voice commands like stop, grip, and release. In the construction industry, speech and natural language-based HRC should be further investigated.

\subsection{Natural language instructions for Non-Construction HRC}
Many studies in which humans give instructions to robots using natural language commands have been conducted for manipulation tasks. Regarding the placing task, Paul et al. \cite{paul2016} and Bisk et al. \cite{bisk2016} leveraged spatial relations in natural language instructions to allow robots to move blocks on the table. Paul et al. \cite{paul2016} proposed a probabilistic model to ground language commands carrying abstract spatial concepts. A neural architecture was suggested for interpreting unrestricted natural language commands in moving blocks identified by a number or symbol \cite{bisk2016}. Mees et al. \cite{mees2020} developed a network to estimate pixelwise placing probability distributions used to find the best placement locations for household objects. However, in order to make a robot perform various construction tasks, it is necessary to use different kinds of attributes describing objects as well as spatial information of the objects.

Several multimodal studies have mapped visual attributes and language information by using two types of input (an image and an instruction). Hatori et al. \cite{hatori} integrated deep learning-based object detection with natural language processing technologies to deal with attributes of household items, such as color, texture, and size. Magassouba et al. \cite{magass2019} proposed a deep neural sequence model to predict a target-source pair in the scene from an instruction sentence for domestic robots. Ishikawa and Sugiura \cite{ishikawa2021} proposed a transformer-based method to model the relationship between everyday objects for object-fetching instructions. 
Guo et al. \cite{guo2023audio} developed an audio-visual fusion framework composed of a visual localization model and a sound recognition model for robotic placing tasks. Murray and Cakmak \cite{murray2022following} and Zhan et al. \cite{zhan2023object} analyzed language instructions about navigation and manipulation tasks to make mobile robots perform various tasks. Murray and Cakmak \cite{murray2022following} proposed a method that uses visible landmarks in search of the objects described by language instructions for household tasks. Zhan et al. \cite{zhan2023object} combined object-aware textual grounding and visual grounding operations for the tasks in real indoor environments. A combination of linguistic knowledge with visual information can describe targets in many ways. The previous studies were intended for robotic household tasks or indoor navigation. 

To utilize these methods for assembly tasks at unstructured and complex construction sites, it is necessary to collect and train construction site images and corresponding language instructions. Previous multimodal studies have relied on thousands to tens of thousands of image-text pairs when training and testing their models. For example, Hatori et al. \cite{hatori} used 91,590 text instructions with 1,180 images, Ishikawa and Sugiura \cite{ishikawa2021} used 1,246 sentences with 570 images, Murray and Cakmak \cite{murray2022following} utilized 25k language data with 428,322 images, and Zhan et al. \cite{zhan2023object} used 90 image scenes with 21,702 language instructions. However, limited image datasets of construction sites present challenges in applying previous multimodal studies of HRC to interactions with construction robots based on natural language instructions.

Some methods interpreted natural language instructions given to robots without relying on visual information. Language understanding using background knowledge \cite{nyga2018} and commonsense reasoning \cite{chen2020} have been explored to infer missing information from incomplete instructions for kitchen tasks. Nyga et al. \cite{nyga2018} generated plans for a high-level task in partially-complete workspaces through a probabilistic model to fill the planning gaps with semantic features. Chen et al. \cite{chen2020} formalized the task of commonsense reasoning as outputting the most proper complete verb-frame by computing scores of candidate verb frames. However, unlike kitchen tasks, it can be challenging to infer targets in construction activities using general knowledge or pre-defined verb frames. Brawer et al. \cite{brawer2018} proposed a model to select one target, described in language instructions, among 20 candidates by contextual information such as the presence of objects and the action history. The context information can also be leveraged in HRC for construction activities, but the proposed model is limited to analyzing language instructions for the pick-up action.

\subsection{Natural language processing in the Construction Industry}
Natural language processing (NLP) is a research domain exploring how computers can be used to interpret and manipulate natural language text or speech \cite{joseph2016}. With the advance of machine learning and deep learning, NLP has been increasingly adopted in the construction industry. NLP applications in construction have been explored in many areas, such as knowledge extraction, question-answering system, factor analysis, and checking \cite{ding2022}. Various documents, such as accident cases \cite{fan2013, kim2019}, injury reports \cite{tixier2016}, compliance checking-related documents \cite{zhang2016}, legal texts \cite{mcgibbney2013}, and construction contracts \cite{lee2020} have been analyzed in construction. Natural language instructions for HRC have not been explored in NLP studies of the construction industry even though HRC through natural language instructions has potential advantages compared with other interfaces such as hand gestures, VR, and brain signals for natural interaction with robots.

Collaboration with a construction robot using natural language instructions requires extracting useful information from the instructions so the robot can start working. In NLP, such information commonly takes the form of entities that carry important meanings as a contiguous sequence of n items from a given text \cite{wu2022}. Previous studies extracted keywords based on frequency features \cite{kosovac2002} and handcrafted rules \cite{liu2021_cost}. These approaches are not robust to unfamiliar input which includes misspelled or unseen words rather than the keywords. To address these challenges, machine learning and deep learning models have been used to extract information about infrastructure disruptions \cite{roy2020} and project constraints \cite{zhong2020, wu2021_info}. However, entities used in these studies, such as task/procedures \cite{wu2021_info}, interval times \cite{zhong2020}, and organization \cite{roy2020} are not suitable for identifying important information from natural language instructions for construction activities. A new group of entities should be defined to give essential information to construction robots.  For example, entities for pick-and-place tasks are relevant to characteristics of the tasks such as target objects, placement location, and placement orientation.

Building Information Modeling (BIM) has been used to visualize and coordinate AEC projects, and can be used for knowledge retrieval since it includes much of the project information \cite{golabchi2013}. The retrieved knowledge from BIM has been applied to plan robot tasks for evaluation of retrofit performance \cite{mantha2018}, indoor wall painting \cite{kim2021_bim} and assembly of wood frames \cite{chong2022}. However, the previous works using BIM information did not consider natural language-based communication with construction robots for HRC. Several studies have used natural language queries to change or retrieve BIM data \cite{lin2016, shin2021, wang2022a}. Lin et al. \cite{lin2016} retrieved wanted BIM information by mapping extracted keywords from queries and IFC entities. However, the proposed method supported only simple queries such as “quantity of beams on the second story” or “quantity of steel columns in the check-in-zone.” Shin and Issa \cite{shin2021} developed a BIM automatic speech recognition (BIMASR) framework to search and manipulate BIM data using a human voice. They conducted two case studies for a building element, a wall, but a quantitative evaluation of the framework was excluded. A question-answering system for BIM consisting of natural language understanding and natural language generation was developed \cite{wang2022a}. Although the system achieved an 81.9 accuracy score with 127 queries, it has a limitation in recognizing complex queries due to rule-based keyword extraction. For example, users can use natural language questions like “What is the height of the second floor?”, “What is the object of door 302?”, or “what is the model creation date?”, which have a similar pattern.

In HRC in construction, robots are expected to perform physical and repetitive tasks as assistants and receive instructions from human workers. Natural language instructions through speech channel can be employed for natural and intuitive interaction. In such scenarios, it is necessary to extract information about construction tasks from the natural language instructions. Previous studies in construction have analyzed text inputs to retrieve useful project information from language queries \cite{lin2016, shin2021, wang2022a}. However, the language queries are different with language commands for construction tasks. There are studies for robot task planning using project information \cite{kim2021_bim, chong2022}, but interaction between human workers and robots were not considered. These studies have limitations in analyzing natural language instructions for robots. There has been no research to plan robot tasks based on natural language commands. To address this research gap, this study proposes a framework for a natural language-enabled HRC method that extracts necessary information from language instructions for robot task planning. In the proposed approach, building component information is used as input to make descriptions of tasks’ attributes more intuitive and simpler.

Table 1 shows the main characteristics of this study. Diverse interaction channels have been considered for interaction with construction robots, but no research investigated how to collaborate with the robots using natural language instructions in construction. This study uses natural language instructions and focuses on pick-and-place operations which are the most common tasks of industrial robots including construction robots. The pick-and-place operation is exploited in many construction tasks like assembly of structural steel elements, bricks, wood frames, tiles, and drywalls by changing types of tools. 
\begin{table}[!hbt]
		\begin{center}
		\caption{Characteristics of this study}
		\label{tab:simParameters}
		\begin{tabular}{|c|c|}
			\hline
			\# & Characteristics      \\ \hline
            1  & \begin{tabular}[c]{@{}l@{}}Communication with construction robots by using natural language \\instructions\end{tabular} \\ \hline
            2  & Pick-and-place operations  \\ \hline
            3  & \begin{tabular}[c]{@{}l@{}}Use of the information of working sites (e.g., designs, materials, ...)\end{tabular}        \\ \hline
            4  & Use of the previous working records   \\ \hline
            5  & \begin{tabular}[c]{@{}l@{}}Target description; ID, dimension, position, and previous records\end{tabular}             \\ \hline
            6  & Case study on drywall installation   \\ \hline
		\end{tabular}
		\end{center}
	\end{table} 
 
While other language instructions used in the previous studies describe target objects and destination, pick-and-place operation for construction activities require one more piece of information about placement orientation. For the variety of patterns, descriptions based on previous working records are also used. As a result, it is required to generate a new dataset for construction activities, and a language model should be proposed and trained. Target objects and destination are described as their IDs, dimension, position or working records available from the construction project information. Thus, this study uses building component information and previous working records to extract essential information that allows the robot to start construction tasks.

\section{System Architecture}
The proposed system aims to make a robot assistant perform construction activities instructed by a human partner using natural language instructions. To this end, a new dataset for pick-and-place construction operations needs to be generated, and a language model trained on this new data should be used, rather than solely relying on the language model used in previous studies. Additionally, the limited availability of image datasets on construction sites can cause difficulties in acquiring surrounding information for robot control. To address this, this study uses building component information available in construction projects to provide robots with the necessary information to execute tasks.

Fig. 1 shows critical components and data workflows of the system, which comprises three modules: Natural language understanding (NLU), Information Mapping (IM), and Robot Control (RC). The NLU module takes a natural language instruction as input and employs a trained language model to perform sequence labeling tasks. Subsequently, the IM module utilizes the output of the NLU and building component information through conditional statements to generate final commands for the RC module. The resulting command is stored in the action history, which serves as one of the inputs to the IM module. Finally, the RC module utilizes three types of information (target, final location, and placement method) to control the robot’s movement for pick-and-place tasks.
\begin{figure}[h!]
    \centering
    \includegraphics[width=.5\textwidth]{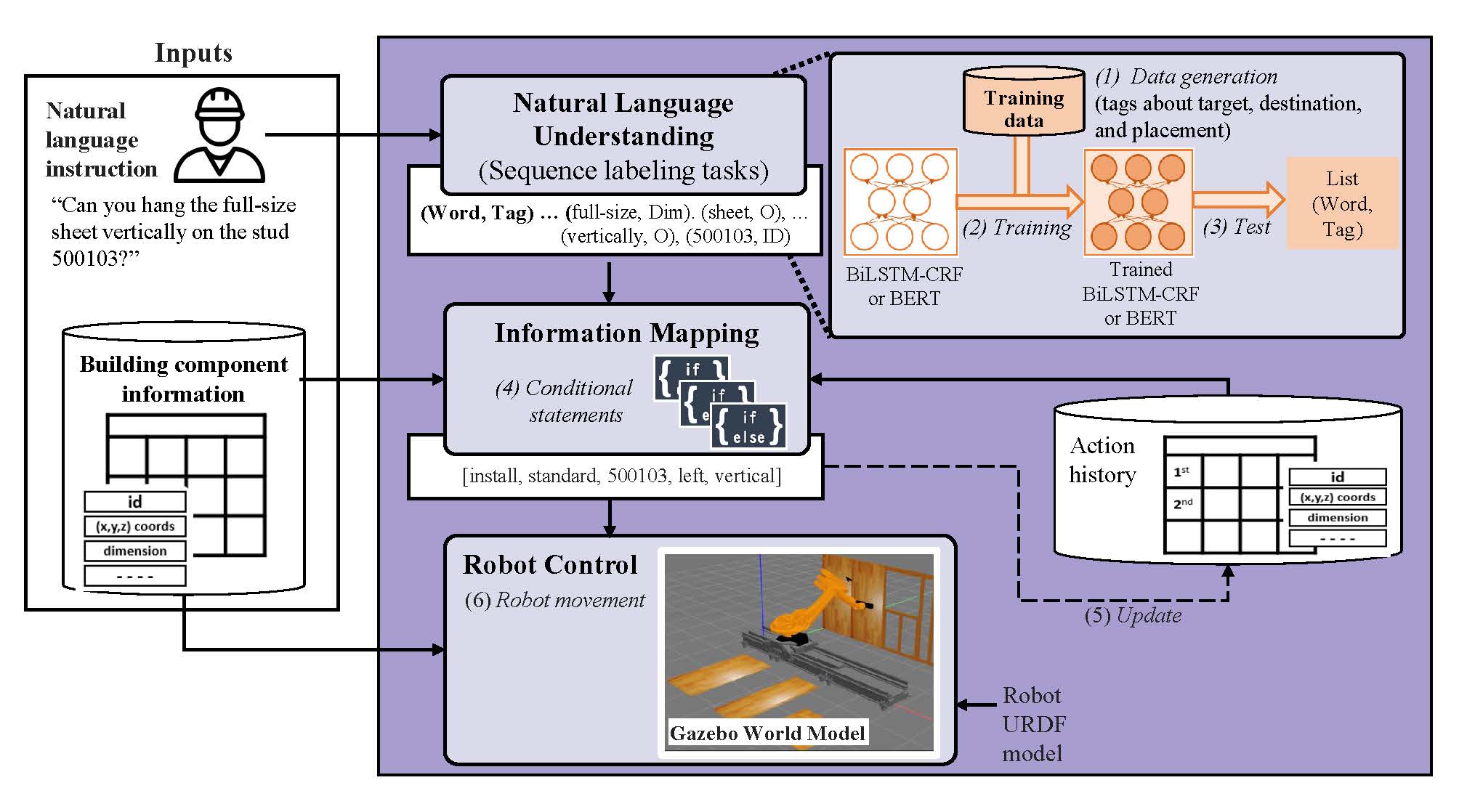}
    \caption{The proposed system using natural language instructions for HRC in construction.}
   \end{figure}

\subsection{Natural Language Understanding (NLU)}
A NLU module aims to predict semantic information from the user’s input which is in natural language. Two main tasks of the NLU are intent classification (IC) predicting the user intent and slot filling extracting relevant slots \cite{benayas2021}. The NLU module of this study focuses on the slot filling which can be framed as a sequence labeling task to extract semantic constituents. Fig. 2 shows an example of the slot filling for the user command “Pick up the full-size drywall to the stud 500107” on a word-level. The word ‘tag’ is used to refer to the semantic label. In this research, two deep learning architectures are utilized to assign appropriate tags to each word of a user command. The first architecture is the Bidirectional Long Short-Term Memory (BiLSTM) layer \cite{graves2005} with a Conditional Random Fields (CRF) layer \cite{lafferty2001}. The second architecture is based on the Bidirectional Encoder Representations from Transformers (BERT) architecture \cite{devlin2018}.
\begin{figure}[h!]
    \centering
    \includegraphics[width=.5\textwidth]{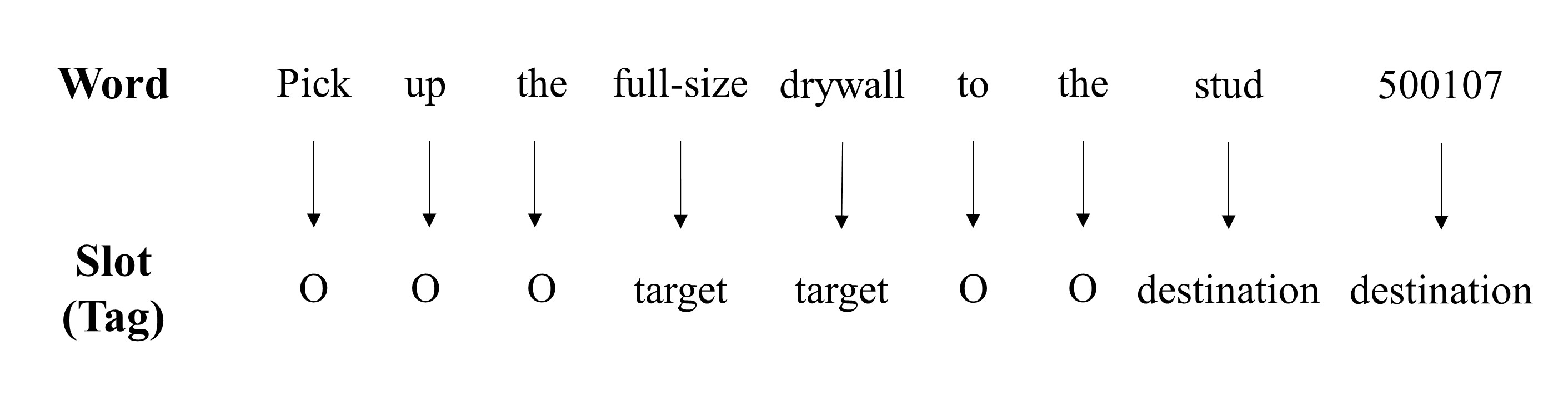}
    \caption{An example of an instruction labeling for slot filling.}
   \end{figure}
   
BiLSTM-CRF is a neural network model that has been used for sequence labeling \cite{huang2015, reimers2017, kong2019}. BiLSTM incorporates a forward LSTM layer and a backward LSTM layer in order to leverage the information from both past and future observations of the sequence. A hidden forward layer is computed based on the previous hidden state (\(\vec{h}_{t-1}\)) and the input at the current position while a hidden backward layer is computed based on the future hidden state (\(\vec{h}_{t+1}\)) and the input at the current position as shown in Fig. 3. At each position t, the hidden states of the forward LSTM (\(\vec{h}_{t}\)) and backward LSTM are concatenated as input to the CRF layer. The CRF layer generates the sequence labeling results by adding some effective constraints between tags. Each tag score output by the BiLSTM is passed into the CRF layer, and the most reasonable sequence path is determined according to the probability distribution matrix. The BiLSTM-CRF model consists of the BiLSTM layer and the CRF layer, which can not only process contextual information, but also consider the dependency relationship between adjacent tags, resulting in higher recognition performance.
\begin{figure}[h!]
    \centering
    \includegraphics[width=.5\textwidth]{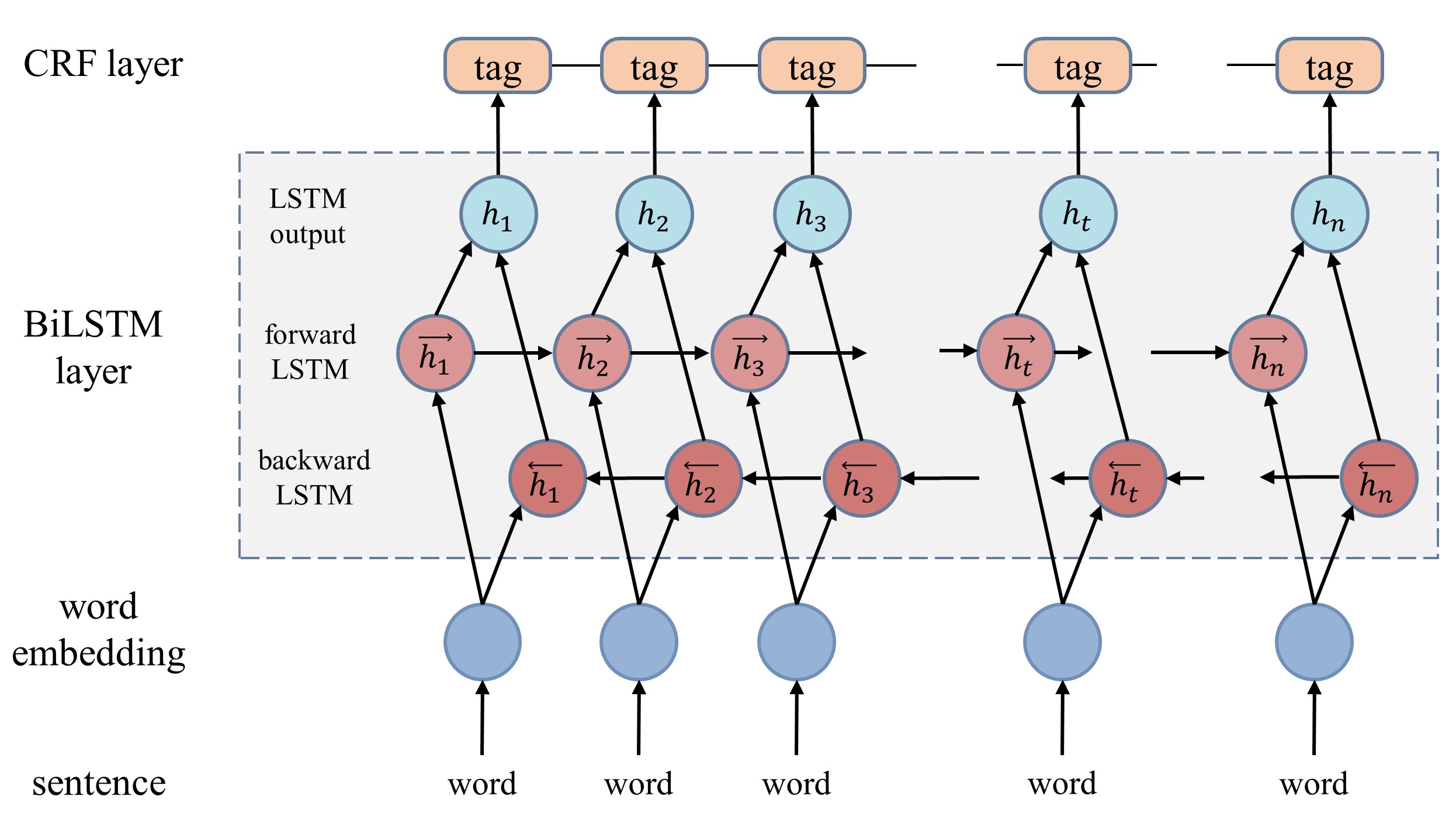}
    \caption{A BiLSTM-CRF structure.}
   \end{figure}
BERT, Bidirectional Encoder Representations from Transformers, is a bidirectional language model that achieves outstanding performance on various NLP tasks \cite{devlin2018}. The architecture of BERT is a multilayer transformer structure which is based on the attention mechanism developed by Vaswani et al. \cite{vaswani2017attention}. BERT is trained to predict words from its left and right contexts using Masked Language Modeling (MLM) \cite{devlin2018} to mask the words to be predicted. The general idea of BERT is to pre-train the model with large-scale dataset, and parameters of the model can be updated for the given tasks during fine-tuning. In this study, pre-trained BERT-base model \cite{devlin2018} is fine-tuned for sentence tagging tasks. As shown in Fig. 4, the input text is tokenized and special token like [CLS], which stands for classification, is added at the beginning. It is needed to create an attention mask. The input for BERT is the masked sequence and the sum of the token and position embeddings (\({E}_{i}\)). Then, the final hidden vector is denoted as T, which is the contextual representation for each token. The token-level classifier is a linear layer using the last state of the sequence as input. In this study, when a word is composed of several tokens and the prediction results of the tokens are different, the class of the word is determined by the token corresponding to more than half of the tokens.
\begin{figure}[h!]
    \centering
    \includegraphics[width=.5\textwidth]{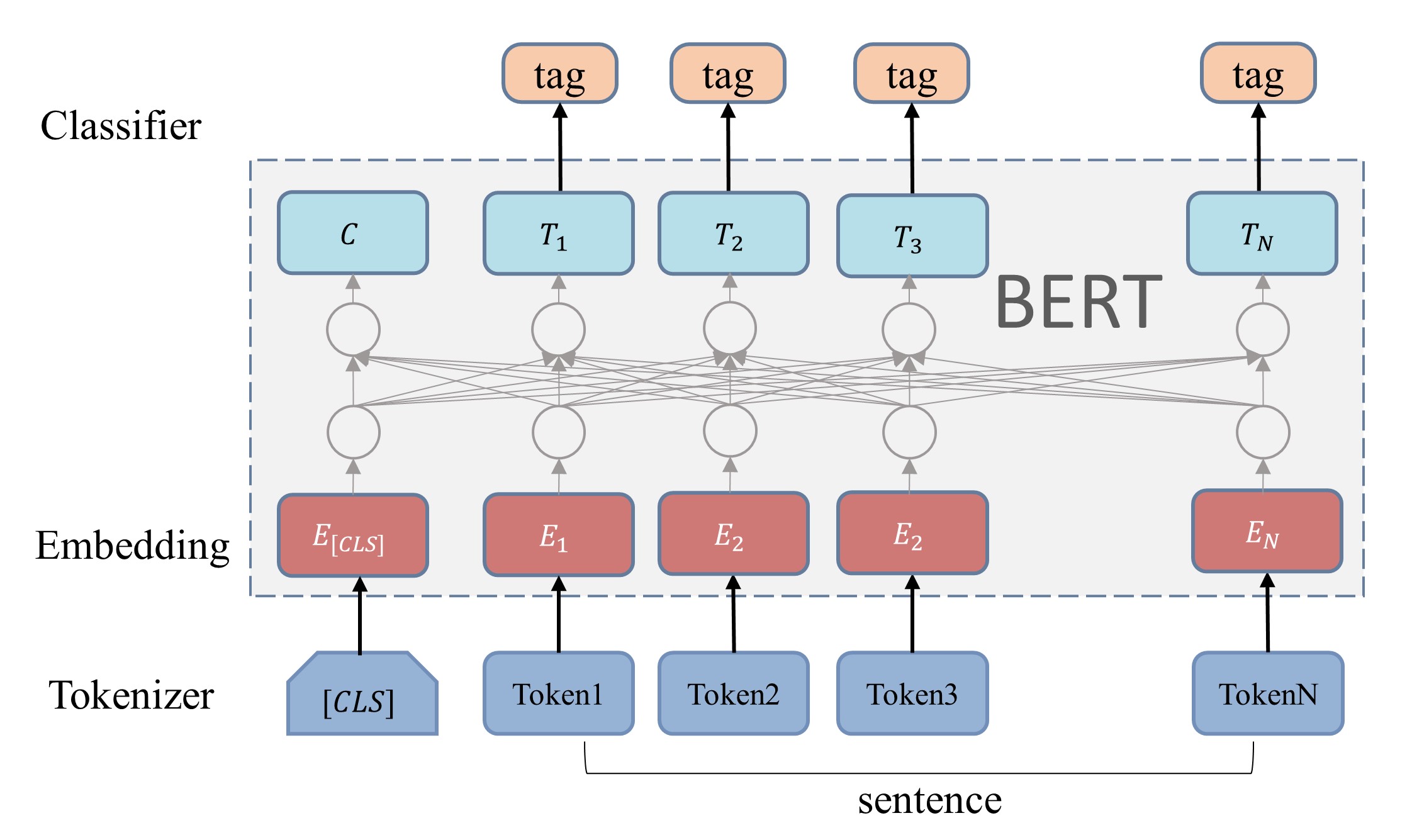}
    \caption{BERT for sentence tagging tasks.}
   \end{figure}
In this study, a language instruction has to deliver one of the characteristics of a workpiece, which can be tags of the language models. In this regard, it is assumed that users have access to mobile devices (e.g., tablet) to obtain building component information such as a name, a unique ID, a dimension, and an initial position of each workpiece on a future construction site. Given the potential use of mobile or wearable technologies in the construction industry \cite{oesterreich2016understanding, jbknowledge, wang2021kamat}, such technologies could be used to provide project information to construction workers making it easier to unambiguously specify which workpieces are to be installed and corresponding location to the robot assistants. 
When generating the language instruction, phrases to describe the three elements should be included and they are tagged as corresponding labels. A target workpiece can be described by its attributes such as name, unique identifier, dimension, and position \cite{akula2011}. A final location is one of the construction workpieces, which is different from the target workpiece. It can be described with its properties. Placement orientation can be expressed as perpendicular, parallel, or other angles. When working records about previous pick-and-place operations are available, a variety of natural language patterns can be utilized in the dataset. For example, the second target workpiece to be installed can be described as “to the left of the previous one” or “same as the previously installed one.”

Information corresponding to the detailed characteristics of the elements is extracted in this NLU module, eliminating the need for additional natural language processing after sequence labeling. Consider the following instruction: "Pick up the panel. The width of the panel is 4 and the length of the panel is 8. Place it vertically to the stud right to the leftmost stud." In this example, the phrase "the width of the panel is 4 and the length of the panel is 8" provides information about the target object while "stud right to the leftmost stud" indicates the destination. In order to align this information with building component information, further natural language processing is required. In this study, language models in the NLU module will be trained with a fine-grained annotated dataset. Consequently, the models can extract corresponding information, including the ID, position, and dimension of the target or destination components.

\subsection{Information Mapping (IM)}
The information mapping module aims to generate a final command for the robotic system using output of the NLU module, building component information, and action history. This module is designed to extract three necessary types of information crucial for a wide range of pick-and-place construction operations, including the identification of a target object, its destination, and placement orientation. The IM module maps NLU output and BIM information and the mapping result is recorded in the action history. The action history record includes information about the last selected object, where the object is placed, and how it is placed. The previous action record can be used to find out the target object and its final location for the current action. The final command to be delivered to the RC module is determined with the latest record of the action history.

To address inconsistencies in the vocabularies between the NLU output and building component information, the module incorporates a procedure that uses conditional statements to extract information about the target object, destination, and placement method. These conditional statements are designed to utilize the ID, position, and dimension information of each component, which can be obtained from the building component information. The appropriate conditional statement to use is determined based on the tag of each word in the NLU output. For instance, if the NLU output contains a tag \emph{ID\_target} that refers to the target object’s ID, the corresponding word is mapped to the ID in the building component information. The component information associated with that ID is then added to the action history as the target object’s information. Similarly, if the NLU output contains a tag \emph{Position\_target} that refers to the position of the target object, the corresponding word is mapped to a component in the building component information within the conditional statement processing the position information. The information associated with that component is then added to the action history.

Various pick-and-place construction operations can be considered for the IM module, including wall tile installation, drywall installation, and bricklaying. For example, in the context of wall tile installation, a command could be "Pick up the 2 by 4 tile and place it horizontally on the lower part of column 300200." In the case of drywall installation, a command could be “Can you hang the panel in the middle to the leftmost stud? Place it to the top part.” Similarly, in bricklaying, a command could be "please put a standard brick vertically next to the previously placed one." These examples highlight the consistent need for precise information about the target object, the destination, and the placement method. The IM module can utilize essential details such as the ID, location, and dimension and generate appropriate commands for the robotic system.

The performance of the IM module is closely linked to the output generated by the Natural Language Understanding (NLU) module, as the latter's output serves as the input for the former. This interdependence implies that the accuracy of the IM module depends on the performance of the NLU module. If there are inaccuracies or misinterpretations in the results predicted by the NLU module, it can lead to errors in the conditional statements of the IM module, hence influencing its operational integrity. Therefore, the accuracy of the IM module is essentially equivalent to the instruction-level accuracy of the NLU module. This relationship underscores the importance of precision and reliability in each component of the system, highlighting the interplay of accuracy across modules.

Once the action history is updated, the final command for robot control is determined as the target object type, destination ID, and placement methods from the action and transferred to the Robotics Control (RC) module. 

\subsection{Robot Control (RC)}
This study uses a virtual robot digital twin to verify that natural language instructions can be used to interact with construction robots in the proposed system. The robot in this study is simulated using ROS (Robot Operating System) and Gazebo that is the virtual environment offered by the Open Source Robotics Foundation. Robot motion planning and execution methods are based on a previous study described in Wang et al. \cite{wang2021kamat}. While Wang et al. \cite{wang2021kamat} used a hand controller to determine the message to be delivered to the robot, this study uses a robotic command generated from the IM module, where the input is natural language instructions. Unlike the previous study, the RC module enables the robot to install the target panel either vertically or horizontally, depending on the placement method information in the input message, and can place it on the middle line or left edge of a stud. The robotic arm movement, which is generated by MoveIt \cite{chitta2012}, has higher priority than the base movement to reduce localization error. When the robot is carrying a target object, collision checking process is applied while the target is considered as part of the robot, so that the robot and the target object will not collide with their surroundings. A human operator can give the next instructions after target placement is completed.

\section{Research Methodology}
A case study is presented for drywall installation to articulate details of the proposed method for natural interaction with robots. For this case study, a 6 degrees-of-freedom KUKA robotic arm is used, and environments for drywall installation are emulated in the Gazebo simulator. The KUKA robot is positioned between a stud wall and drywall panels and the base of the robot can move in a straight line as shown in Fig. 5(a). The stud wall consists of thirteen studs as illustrated in Fig. 5(b). In this case study, one stud is designated as the final location for place operation and the left edge of a drywall panel is laid on the stud. In general, drywall panels are available in rectangular shapes. Standard panel size is 4 feet wide and 8 feet long and panels of different sizes are cut according to the designed dimensions in construction practice. We use three sizes of panels including the standard ones as well as two unique panel sizes (Fig. 5(c)). The drywall panels will be installed from left to right along the stud wall.
\begin{figure}[h!]
    \centering
    \includegraphics[width=.5\textwidth]{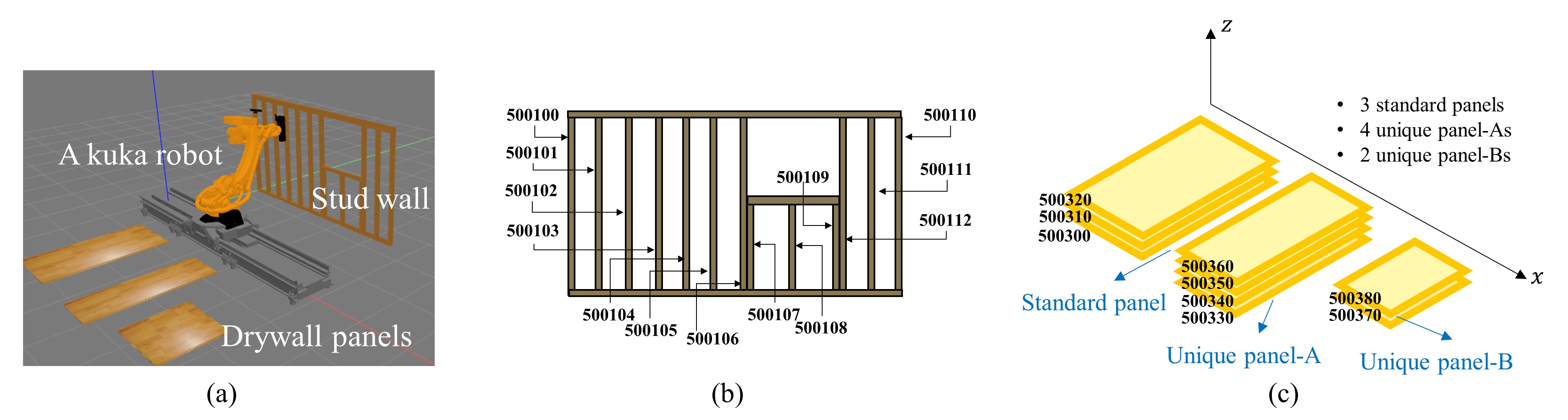}
    \caption{Case study settings for drywall installation: (a) robot operation environment; (b) a stud wall consisting of 13 studs; (c) 9 drywall panels on the floor.}
   \end{figure}
The drywall panels can be installed in a vertical or horizontal orientation. Fig. 6 shows examples of how to place drywall panels onto the studs. Examples of vertical placement are shown in Fig. 6(a), and the left edge of the panel can be placed on the center line of a stud or the left side of a stud. When the panels are placed horizontally perpendicular to studs, they can be placed on the top or bottom part of the studs as shown in Fig. 6(b). Therefore, natural language instructions for drywall placement should include how (i.e., in what configuration) to place the drywall panels.
\begin{figure}[h!]
    \centering
    \includegraphics[width=.5\textwidth]{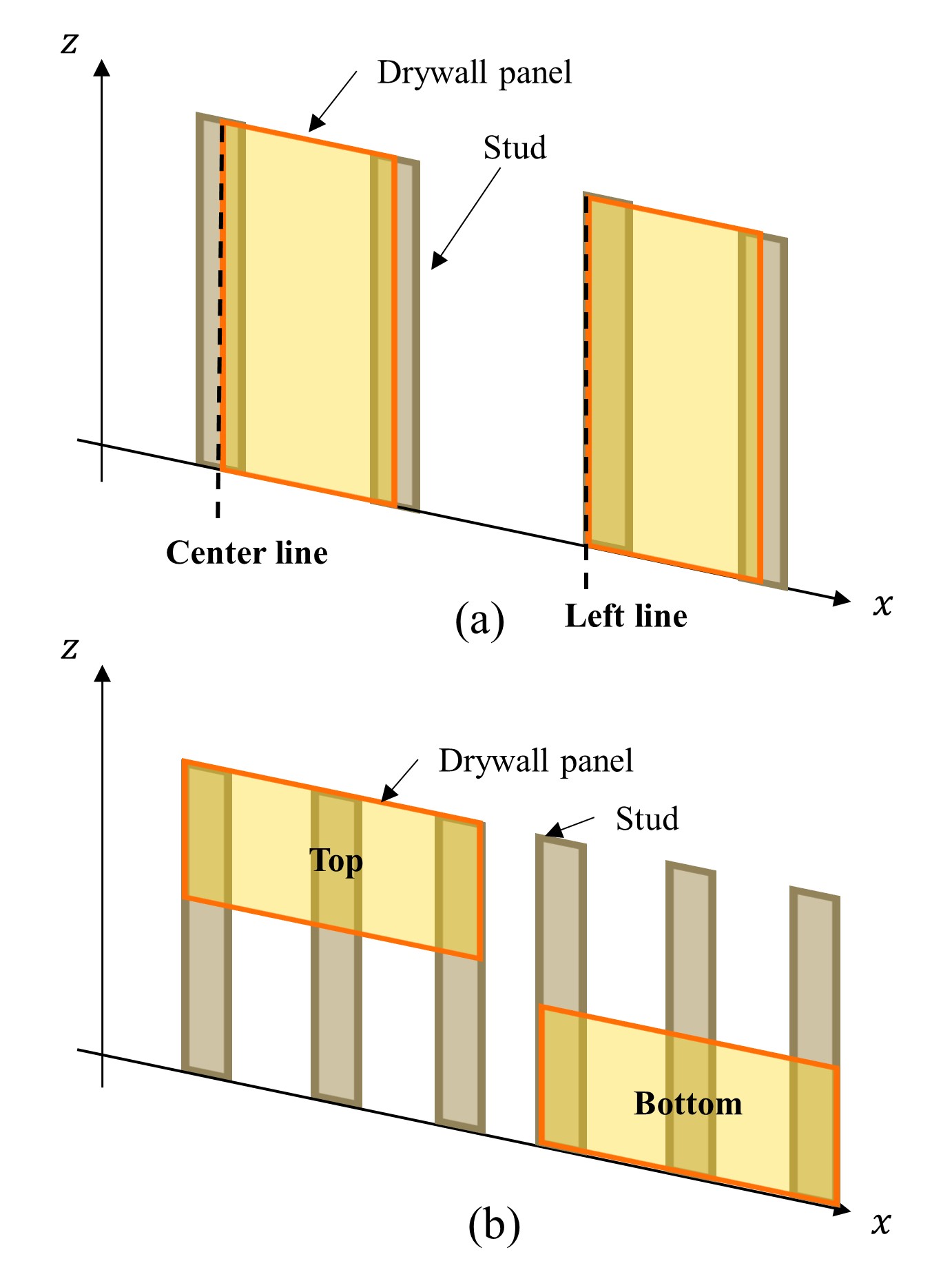}
    \caption{Two ways of drywall installation: (a) vertical placement of drywall panels; (b) horizontal placement of drywall panels.}
   \end{figure}

\subsection{Data Generation and Natural Language Understanding (NLU)}
A new dataset of natural language instructions for drywall installation was created and annotated. Each instruction, consisting of one or multiple sentences, clarifies a desired drywall as a target, a stud as a final location, and how to place the drywall panel. To achieve a fine-grained annotation, this study utilized 12 tags that enabled the classification of these three essential categories into more detailed categories. These tags include six that describe the characteristics of the target object, three that illustrate the final location, and the remaining three for the placement orientation. Each instruction contains these three pieces of information exactly once. In the dataset, there are co-reference issues, where words referring to a target object, a final location, and a placement method can be included multiple times within a single instruction. However, expressions clearly indicating features related to these three types of information appear only once in each instruction. 
The final location and the target are one of the building components illustrated in the Fig. 5(b) and Fig. 5(c), respectively. To utilize widely used expressions in language instructions, construction videos about drywall installation \cite{youtube} and other studies exploring pick-and-place language instructions were considered when generating the new dataset. In these language instructions, drywalls and studs are described by combinations of representations related to ID, dimensions, and relative location. 

A drywall panel is represented by its ID, dimension, or position, while a stud is represented by its ID or position (Fig. 7). BIM models used in previous studies have allocated a five to seven- digit number to every building element \cite{liu2015, heigermoser2019, fazeli2021}. Each element ID is represented as a unique 6\-digit number in this case study and is tagged with \emph{ID\_stud} and \emph{ID\_wall} for stud and a drywall panel, respectively. A list of digits can be read out in the working environments such as warehouses or factories to increase work performance \cite{berger2007reducing, goomas2010ergonomics, goomas2022increasing}. While it may not be common to utter long digits in today’s construction workers’ practice, this study suggests that using IDs could be one of the effective ways for workers to unambiguously indicate a target object when interacting with robots to ensure accurate selection and installation of workpieces. 
\begin{figure}[h!]
    \centering
    \includegraphics[width=.5\textwidth]{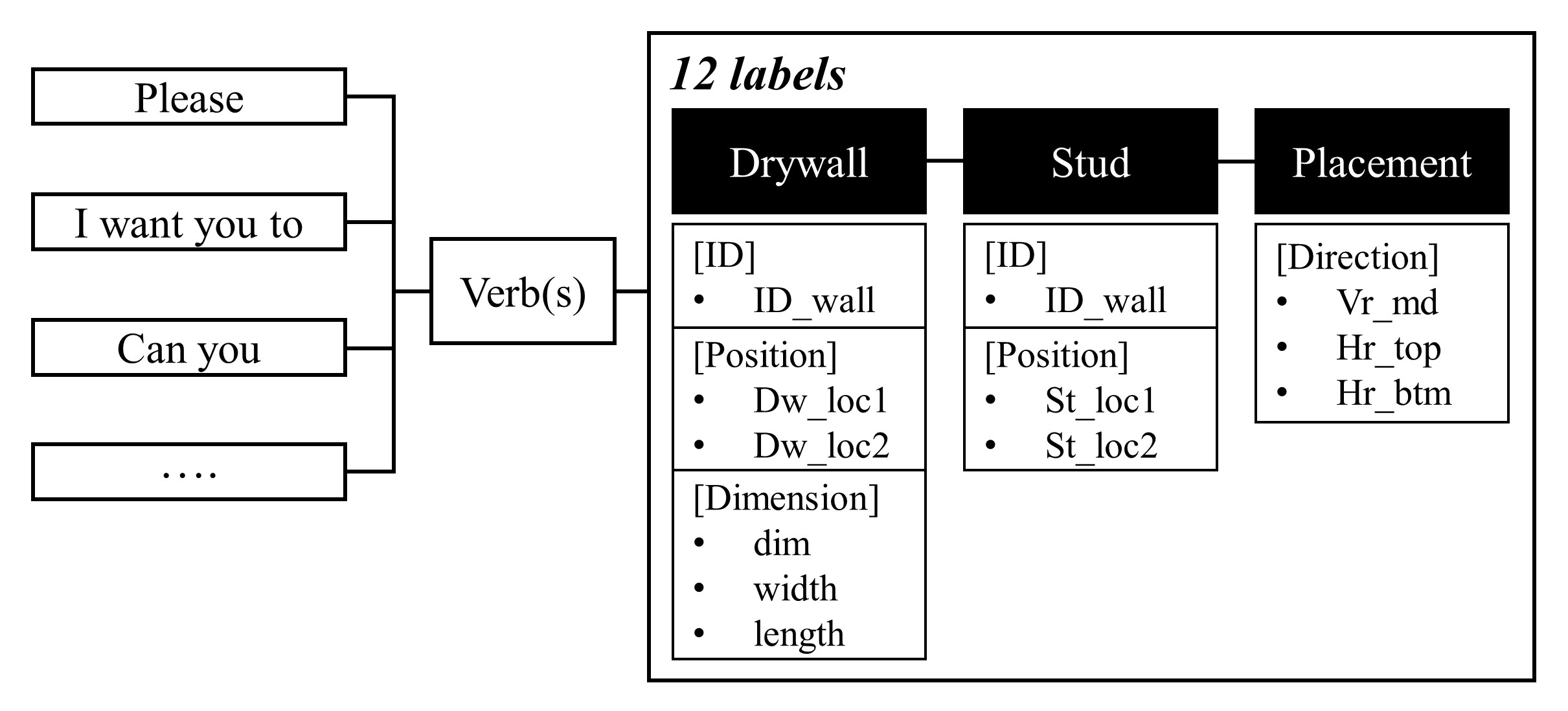}
    \caption{Dataset generation for drywall installation.}
   \end{figure}
   
The dimensions of the target drywalls are labeled with length, width, or dim. When a target object is described in numbers such as “4 by 8” or “the length is 8”, the numbers are annotated as length or width. Dimension of the target object can be expressed with words “full-size”, “standard”, or “full”, and the words representing the size of the target object are annotated as label dim. 

Both a target drywall panel and a final location (stud) can be described as their locations using one perspective view in this case study. For example, stud 500100 is the leftmost stud and drywall sheets 500300, 500310, and 500320 are the leftmost ones as shown in Fig. 5. The words to indicate locations of the stud and drywall panels are labeled as \emph{St\_loc1} and \emph{Dw\_loc1}. When describing the locations, the relationship of a place to other places can be used. It means that the location changes based on the secondary location. When a final location of stud is described using relative location, both \emph{St\_loc1} and \emph{St\_loc2} are used together while both \emph{Dw\_loc1} and \emph{Dw\_loc2} are used together when the target drywall is described. For example, in Fig. 5, the location of the stud 500101 can be expressed as “second left to the stud 500103” or “right to the stud 500100.” In this case, the direction like “second left” or “right” is also annotated as \emph{St\_loc1} and the word “500103” or “500100”, which is corresponding to the secondary location, is annotated as \emph{St\_loc2}.

Finally, regarding how to place drywall panels, there are three labels of \emph{Vr\_md}, \emph{Hr\_top}, and \emph{Hr\_btm}. When a panel is vertically placed on the middle line of the stud, the corresponding words like “middle line” or “center line” is labeled as \emph{Vr\_md}. When a target object is placed horizontally on the top row of a stud or on the bottom row of a stud, the corresponding words are annotated as \emph{Hr\_top} or \emph{Hr\_btm}. Terms like “upper part”, “upper horizontal row”, and “top part” are annotated as \emph{Hr\_top} while terms like “lower part” and “bottom row” are annotated as \emph{Hr\_btm}. Given this variability, the same words should be annotated as different tags, creating a challenge for language models to correctly interpret the intended context. When a placement method is not mentioned in a language instruction, it means that the panel is installed vertically on the left line of the stud. It is considered default in this study and the language instruction does not have a tag about this placement method. 

There are a total of 13 labels, with 12 of them representing either a target drywall, a final location (stud), or a placement method, as shown in Fig. 7. The remaining label, referred to as ‘O’, is utilized to signify that the corresponding word is not associated with any entity. If a target, a destination, or a placement is mentioned multiple times in a single instruction, words that do not deliver any characteristics of the three information are tagged as ‘O.’ For example, in a three-sentences instruction “Please move the drywall board and drive it vertically in the center line of the stud. The width is 4 and the length is 8. The stud is laying on the left to the 500103”, ‘the drywall board’ and ‘it’ in the first sentence refer to a target object but they do not deliver any important characteristic, so they are tagged as ‘O.’

In total, 1,584 natural language instructions with the 13 labels for drywall installation were generated and manually annotated. These instructions consist of 3,072 sentences and a total word count of 39,841. The dataset was split into three parts: 1,268 instructions for training (80\%), 158 instructions for validation (10\%), and 158 instructions for test (10\%). Table 2 shows annotation results of the 1,584 instructions. The dataset includes fine-grained details of the target objects, expressed through six tags: \emph{Dw\_loc1}, \emph{Dw\_loc2}, \emph{ID\_wall}, \emph{dim}, \emph{length}, and \emph{width}, which account for a total of 2,535 words. Similarly, the destination details are captured using the tags \emph{ID\_stud}, \emph{St\_loc1}, and \emph{St\_loc2}, encompassing 4,166 words. Additionally, the dataset incorporates placement orientation information, classified into three distinct classes, and comprising a total of 2,060 words. Consider the example instruction: " Can you install the piece 500310 vertically in the stud? The stud is laying third to the left from the stud 500105. Please hang the panel into the middle line." This approach allows for extraction of specific details, such as the \emph{ID\_wall} tag for the target, \emph{Dw\_loc1} and \emph{Dw\_loc2} tags for the destination, and the \emph{Vr\_md} tag representing a specific placement orientation rather than simply highlighting three main categories. Such granularity can significantly enhance the richness and precision of the data interpretation.
\begin{table}[]
\caption{Annotation results of the dataset}
\centering
\begin{tabular}{|l|l|}
\hline
Tags & Number of words \\ \hline
\textit{Dw\_loc1} & 702 \\ \hline
Dw\_loc2 & 368 \\ \hline
\textit{Hr\_btm} & 184 \\ \hline
\textit{Hr\_top} & 210 \\ \hline
\textit{ID\_stud} & 550 \\ \hline
\textit{ID\_wall} & 514 \\ \hline
\textit{O} & 31,080 \\ \hline
\textit{St\_loc1} & 2,652 \\ \hline
\textit{St\_loc2} & 964 \\ \hline
\textit{Vr\_md} & 1,666 \\ \hline
\textit{dim} & 259 \\ \hline
\textit{length} & 346 \\ \hline
\textit{width} & 346 \\ \hline
SUM & 39,841 \\ \hline
\end{tabular}
\end{table}

While the first author performed the initial manual annotation, two other individuals checked the appropriateness of annotation guidelines by annotating the test dataset in two rounds. Appendix I presents the annotation guidelines used in this study. In the first round, the two annotators labeled the dataset based on the annotation guidelines and several examples. The annotators achieved 96.05\% and 89.24\% accuracy, respectively. They received feedback on the results of the first-round annotation. In the second round, both annotators achieved 98.15\% and 98.56\% accuracy in annotation, which are almost 100\% accuracy. Any errors in the second round were simple human errors. The validation set is used to compare the performance of different models in the NLU module. The model with the best performance on the validation dataset is used to evaluate the test dataset and the results are delivered to the IM.

The specific parameters of the BiLSTM-CRF model used in this case study are determined based on previous studies \cite{huang2015, reimers2017, tang2020} as follows: the number of neural network layers is 2; word embedding size is 50; the number of hidden layer LSTM neurons is 300; batch-size is 16; the dropout is 0.1; the optimizer is set to Adam \cite{kingma2014} with a learning rate of 0.001; the Adam optimizer trains 20 epochs. The total number of parameters is about 250,000. In the case of BERT, “BertForTokenClassification” class was used to find-tune the BERT-base-uncased model of the original BERT \cite{devlin2018}. The specific parameters are as follows: the number of encoder layers is 12; the number of attention-heads is 12; the number of hidden units: 768; batch-size is 16; the dropout is 0.1; the optimizer is Adam with a learning rate of 3e-5; the number of training epochs is 5. The total number of parameters is 110 million. Fig. 8 shows network architecture diagrams of BiLSTM-CRF and BERT.
\begin{figure}[h!]
    \centering
    \includegraphics[width=.5\textwidth]{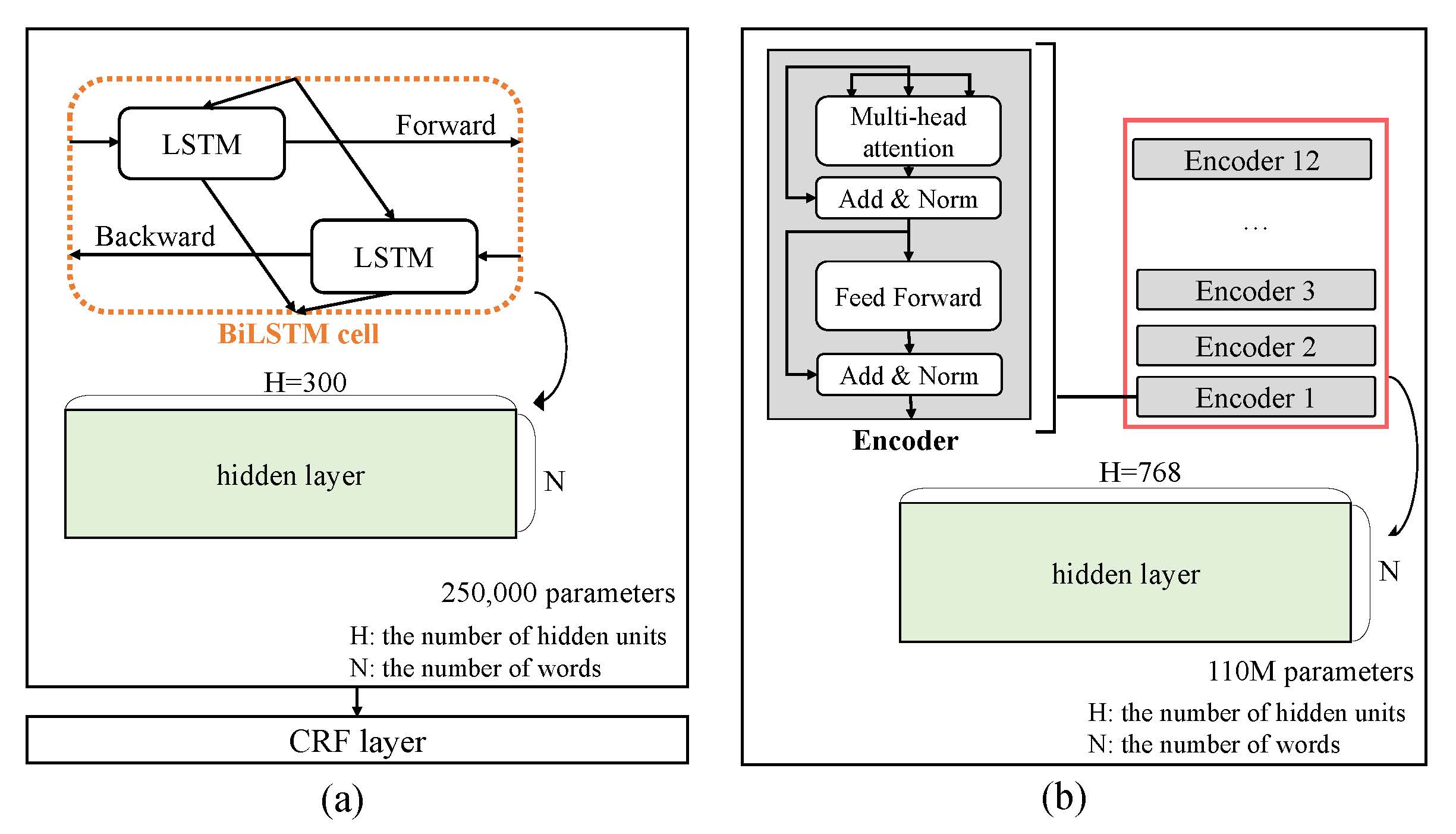}
    \caption{Network architecture diagrams: (a) BiLSTM-CRF; (b) BERT.}
   \end{figure}
\subsection{Information Mapping (IM)}
The IM module utilized several rules to extract final information about a target panel, a stud as destination, and a placement method based on the output of the NLU module and building component information (Fig. 9). The output of this module is recorded in an action history table as nine types of values: \emph{stud\_id} (ID of the stud), \emph{installed\_x\_left} (x coordinate of the left side of the installed panel), \emph{installed\_x\_right} (x coordinate of the right side of the installed panel), \emph{left\_cent} (if the panel is installed on the left side of the stud or the center line of the stud), \emph{ver\_hor} (if the panel is installed vertically or horizontally), \emph{top\_btm} (if the panel is installed on the top row or the bottom row), \emph{drywall\_id} (ID of the drywall panel), \emph{w} (width of the drywall panel), and \emph{l} (length of the drywall panel). The records in the action history table can be used to extract the final command for the robot control.

\begin{figure}[h!]
    \centering
    \includegraphics[width=.5\textwidth]{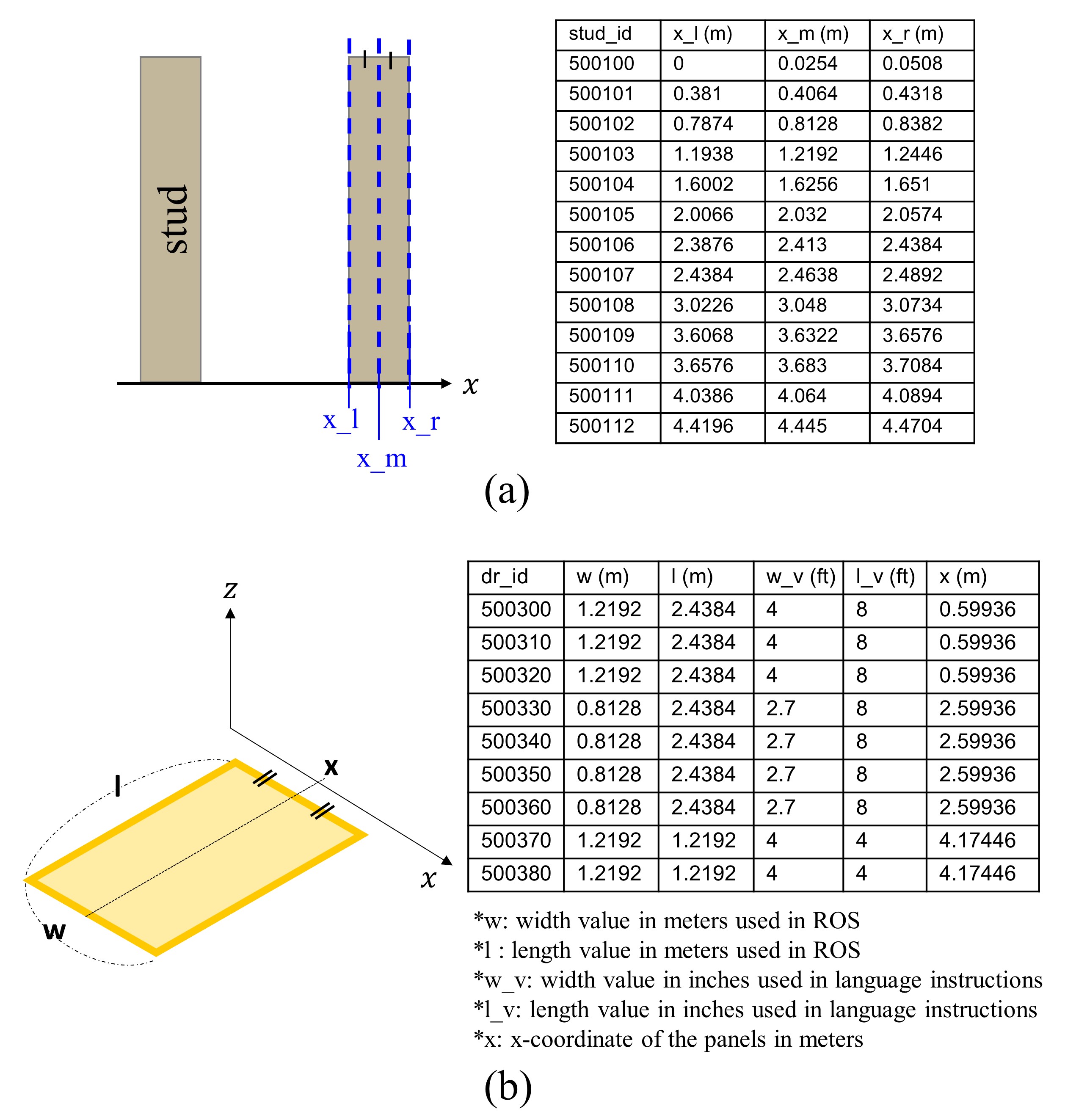}
    \caption{Stud and drywall information. (a) x-coordinates of the thirteen studs; (b) dimensions and x-coordinates of the nine drywall panels.}
   \end{figure}
   
The rules of the IM module about drywall panels are shown in Figs. 10 and 11. The pseudocode in Fig. 9 can be used when a target of pick-and-place operation is described as its dimension. If the dimension of the target drywall panel is described by its length and width values or words like ‘standard’ and ‘full-size’, the target features are extracted by its length and width values in the drywall information table in Fig. 9(b), which is marked as \emph{TableD} in Fig. 10. When an expression for a previously performed operation is used, such as “previously installed”, the target of the last performed operation is retrieved from the action history table \emph{ActHist} and the panel with the same characteristics is determined as the target of the current operation.
\begin{figure}[h!]
    \centering
    \includegraphics[width=.5\textwidth]{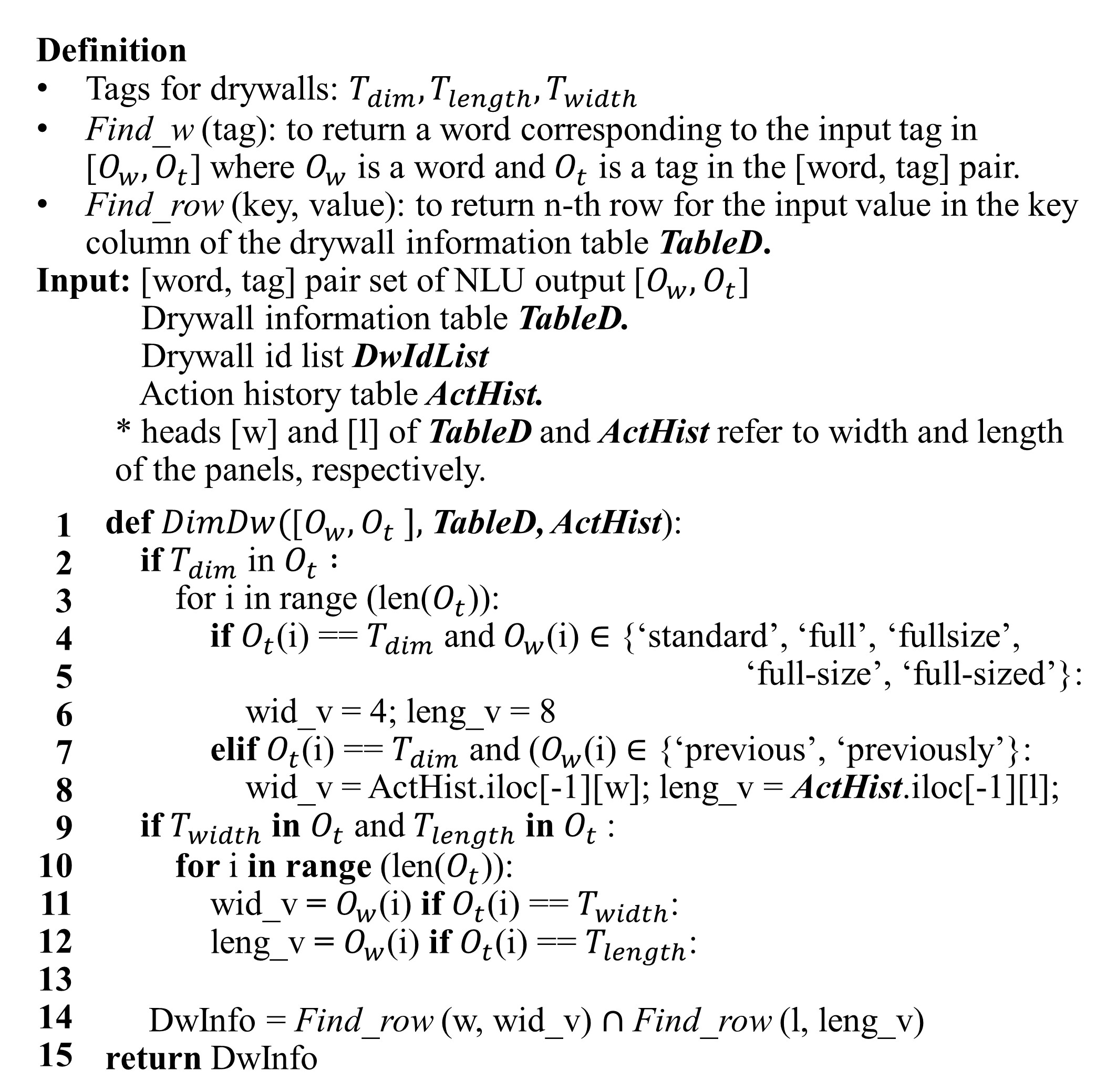}
    \caption{Pseudocode for information extraction about drywall panels using dimension-related tags.}
   \end{figure}
Fig. 11 shows pseudocode for the process used when drywall panels are labeled as their IDs or position. When the tag of \emph{ID\_wall} is included in the output of the NLU, the information of the panel corresponding to that tag is returned. If only \emph{Dw\_loc1} refers to a workpiece at the output of the NLU module, the target is determined by the x coordinate value for the initial position of drywall panels and the word tag to \emph{Dw\_loc1}. In the case that both of \emph{Dw\_loc1} and \emph{Dw\_loc2} are included in the output of NLU, a target panel is explained by its relative location that changes based on the secondary location. The x coordinate of the target panel’s initial position, which is finally used to extract the target information, is determined from the secondary place and the direction tagged with \emph{Dw\_loc2} and \emph{Dw\_loc1}, respectively.
\begin{figure}[h!]
    \centering
    \includegraphics[width=.5\textwidth]{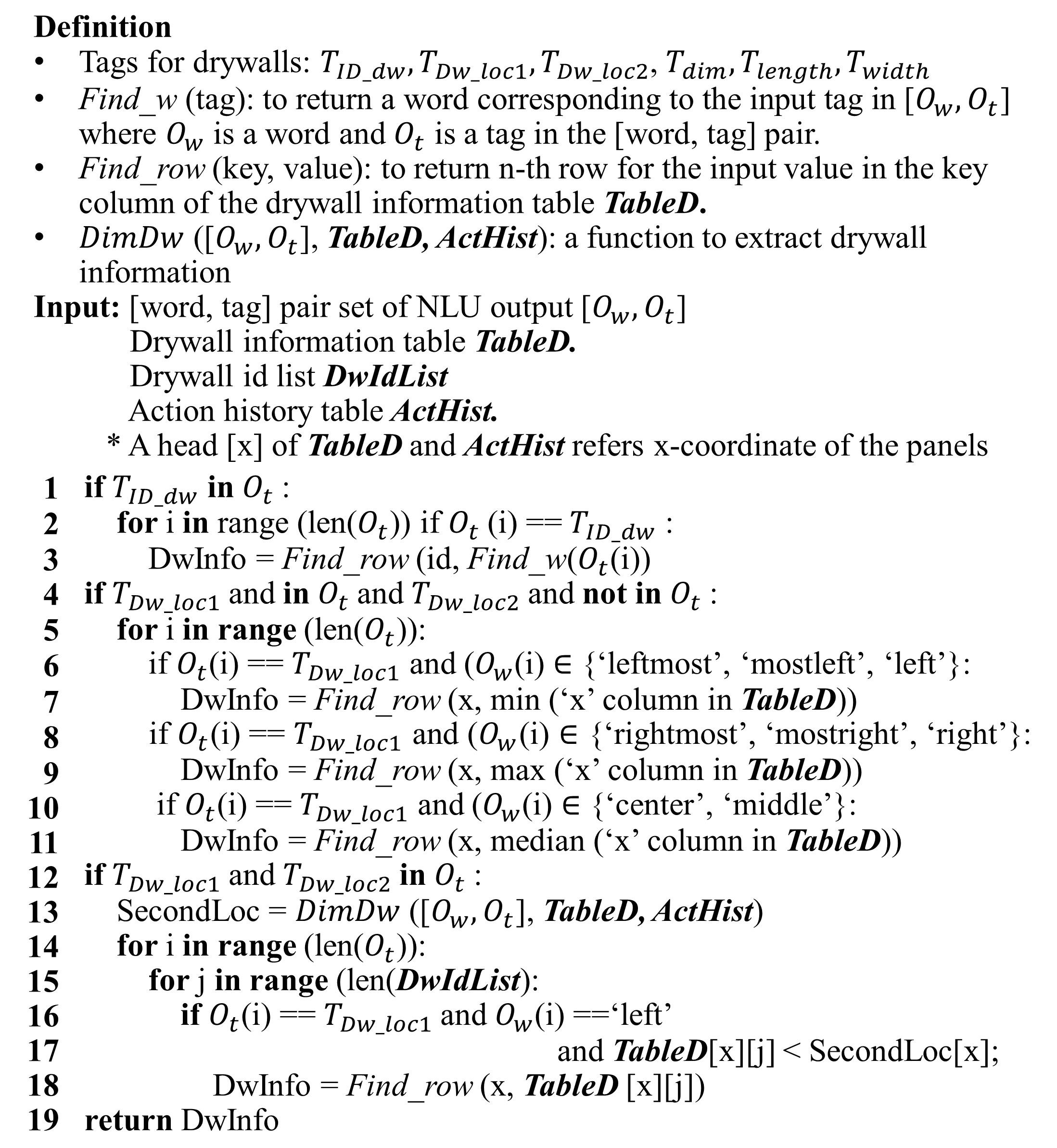}
    \caption{Pseudocode for information extraction about drywall panels using tags of ID and positions.}
   \end{figure}
Fig. 12 shows how to extract information for a stud that is a final location for pick-and-place operations. When the tag of \emph{ID\_stud} is included in the output of the NLU, the information of the stud corresponding to that tag is returned. Otherwise, the output of NLU includes \emph{St\_loc1} or \emph{St\_loc2}, so that the stud is described by its location. When \emph{St\_loc2} is not included, the stud is either the leftmost one or rightmost one. When both \emph{St\_loc1} and \emph{St\_loc2} are extracted, the stud as final location is determined by the spatial relationship described by words tagged by \emph{St\_loc1} and \emph{St\_loc2}.
\begin{figure}[h!]
    \centering
    \includegraphics[width=.5\textwidth]{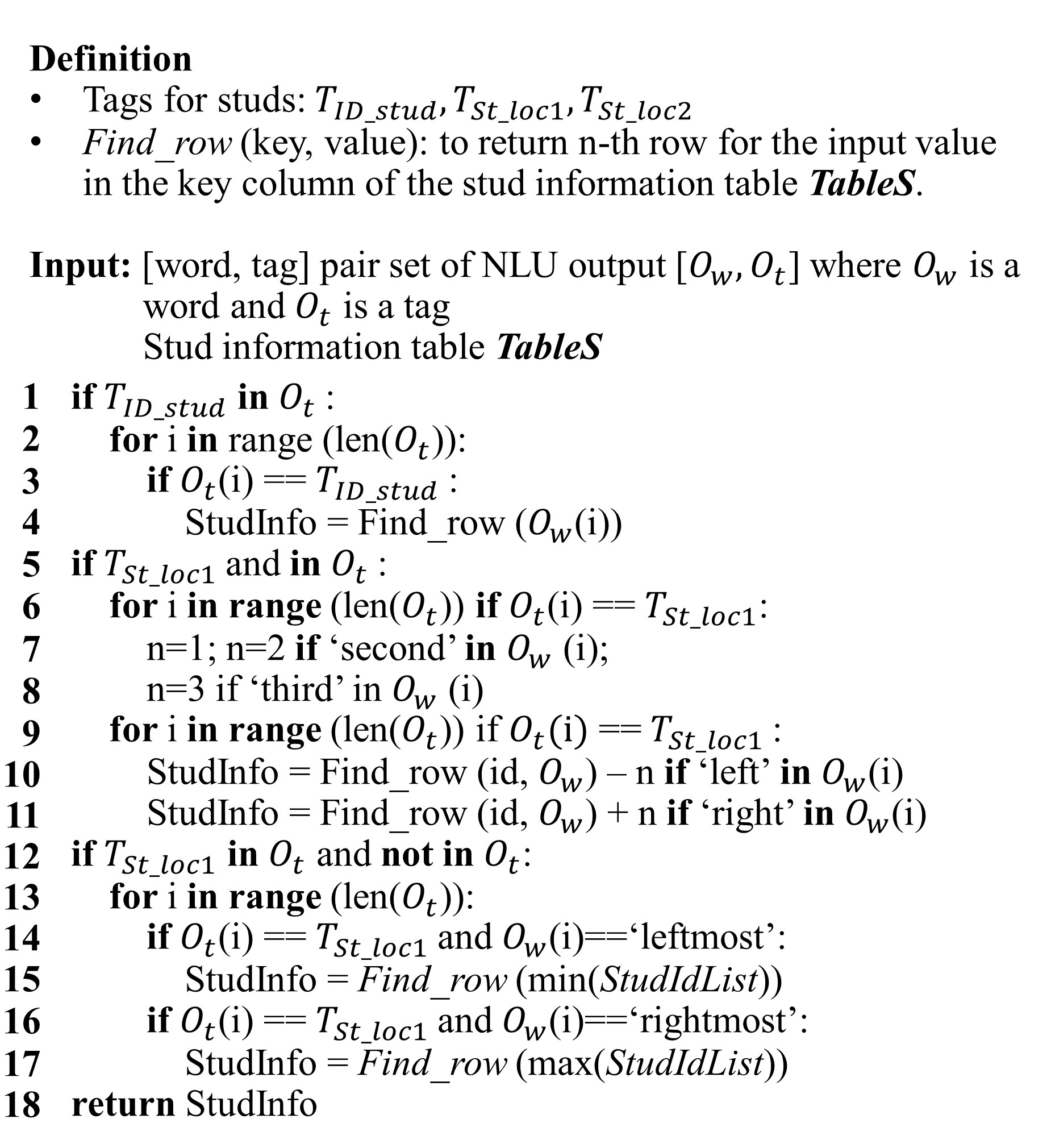}
    \caption{Pseudocode for information extraction about studs.}
   \end{figure}
To start a pick-and-place operation for drywall installation, it is essential to know the placement method as well as the target and final location. Three types of placement methods are used in this study: \emph{Vr\_md}, \emph{Hr\_top}, and \emph{Hr\_btm}. If the output of the NLU module does not contain these three tags, the left edge of the drywall panel is set to be placed vertically to the left of the stud. The three pieces of information about the current job are recorded in the action history table. The \emph{installed\_x\_left} value in the action history table is determined according to the combination of the placement method and the final location, and the \emph{installed\_x\_right} value is calculated based on the placement method, the target, and the \emph{installed\_x\_left} value.

\section{Experimental Results}
This study trained the BiLSTM-CRF model and BERT by varying the number of training data to see the effects of training data size on the performance of the model. With different amounts of training data, four models with the same architecture were trained for both language models. Fig. 13(a) reports the training accuracy of the four BiLSTM-CRF models across the 20 epochs. The four BERT models were trained across the 5 epochs since they converged quickly as shown in Fig. 13(b). The accuracy of the LSTM-M1 and BERT-M1, which were trained with ample training data, showed a considerably faster increase in the learning progress early in training.

\begin{figure}[h!]
    \centering
    \includegraphics[width=.5\textwidth]{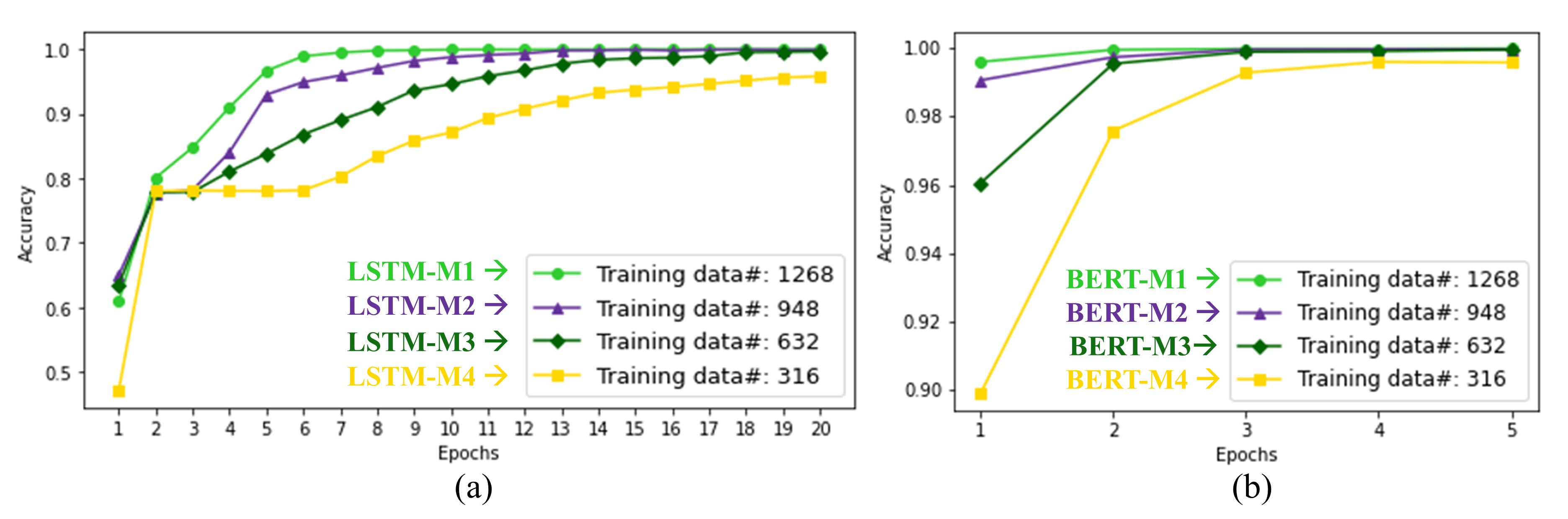}
    \caption{Comparison of training accuracy: (a) BiLSTM-CRF and (b) BERT.}
   \end{figure}
The performance of the eight models were evaluated on the validation set and compared in Table 3. In this study, two types of accuracy are computed to measure performance. Word-level accuracy (\emph{Acc\_word}) was computed based on the number of all the words in the dataset, which provides the proportion of words that are correctly predicted. The eight models achieved high \emph{Acc\_word} over 96\%. However, even one tag incorrectly predicted in a language command can affect the IM module that derives the final robot command, causing disruptions in the robot’s performance. To address this problem, Instruction-level accuracy (\emph{Acc\_inst}) considers whether all words in each instruction are correctly predicted or not, thus providing the proportion of language instructions in which all words are correctly predicted. For example, as shown in Table 3, \emph{Acc\_word} of LSTM-M4 was measured as high as 96.13\%, but \emph{Acc\_inst} of LSTM-M4 showed an accuracy of 48.73\%. This means that the robot can accurately perform 48\% of the given language instructions.
\begin{table}[]
\centering
\caption{Comparison of model performance on validation dataset.}
\begin{tabular}{|l|ll|ll|}
\hline
\multirow{2}{*}{Model} & \multicolumn{2}{l|}{Result 1} & \multicolumn{2}{l|}{Result 2} \\ \cline{2-5} 
 & \multicolumn{1}{l|}{$N_{w}$} & \emph{Acc\_word} & \multicolumn{1}{l|}{$N_{l}$} & \emph{Acc\_inst} \\ \hline
LSTM-M1 & \multicolumn{1}{l|}{2} & 99.95\% & \multicolumn{1}{l|}{1} & 99.37\% \\ \hline
LSTM-M2 & \multicolumn{1}{l|}{2} & 99.95\% & \multicolumn{1}{l|}{2} & 98.73\% \\ \hline
LSTM-M3 & \multicolumn{1}{l|}{11} & 99.73\% & \multicolumn{1}{l|}{9} & 94.30\% \\ \hline
LSTM-M4 & \multicolumn{1}{l|}{144} & 96.13\% & \multicolumn{1}{l|}{81} & 48.73\% \\ \hline
BERT-M1 & \multicolumn{1}{l|}{0} & 100.00\% & \multicolumn{1}{l|}{0} & 100.00\% \\ \hline
BERT-M2 & \multicolumn{1}{l|}{1} & 99.97\% & \multicolumn{1}{l|}{1} & 99.36\% \\ \hline
BERT-M3 & \multicolumn{1}{l|}{6} & 99.85\% & \multicolumn{1}{l|}{6} & 96.20\% \\ \hline
BERT-M4 & \multicolumn{1}{l|}{43} & 98.90\% & \multicolumn{1}{l|}{33} & 79.11\% \\ \hline
\end{tabular}
{\small
        \begin{tablenotes}
        \item[a] $N_{w}=$ the number of incorrect prediction of words.
        \item[b] \emph{Acc\_word} $=(3895-N_{w})/3895$
        \item[c] $N_{l}=$ the number of language instructions including incorrect prediction
        \item[d] \emph{Acc\_inst} $=(158-N_{l})/158$
        \end{tablenotes}
        }
\end{table}
Out of all eight models, BERT-M1 achieved the highest accuracy, with 100.00\% accuracy at both the word-level and instruction-level. Generally, model performance increased with larger amounts of training data. BERT models, including BERT-M1, outperformed the BiLSTM-CRF model when trained on equivalent amounts of data. Even with a small dataset (BERT-M4), the model achieved an instruction-level accuracy of 79.11\%, demonstrating the effectiveness of fine-tuning pre-trained models in such cases. The study also confirmed that training with a minimal amount of data (equivalent to twice the validation set) resulted in a rapid decline in accuracy compared to the other models.

The number of false predictions for the 13 tags is compared in Table 4. LSTM-M1 and LSTM-M2 had two wrong predictions for \emph{Dw\_loc1} and \emph{Vr\_md}, respectively. As in the example in Fig. 14(a), ‘most left’ was incorrectly predicted as \emph{St\_loc1} representing a stud instead of \emph{Dw\_loc1} representing a drywall panel. Within our dataset, the word ‘middle’ is contextually labeled as \emph{Vr\_md} or \emph{Dw\_loc1}, which can occasionally increase the complexity of predictions. Fig. 14(b) shows that the word ‘middle’ was predicted as \emph{Dw\_loc1} instead of \emph{Vr\_md} indicating the placement method. BERT-M2 also had one error, the word ‘middle’ corresponding to \emph{Dw\_loc1} was predicted as \emph{Vr\_md} (Fig. 14(c)). These results may be due to the similarity of the words referring to the position and the placement method. Such issues tend to be mitigated when language models are trained with a large amount of data as shown in the previous deep learning-based studies \cite{wei2018empirical, mathew2021deep}.
\begin{table*}[t]
\caption{Comparison of incorrect prediction of each class for the four models.}
\centering
\begin{tabular}{|l|c|cccccccc|}
\hline
\multicolumn{1}{|c|}{\multirow{2}{*}{Tags}} &
  \multicolumn{1}{l|}{\multirow{2}{*}{\begin{tabular}[c]{@{}l@{}}\# of words\\ (Ground\\ truth)\end{tabular}}} &
  \multicolumn{8}{c|}{Incorrect prediction} \\ \cline{3-10} 
\multicolumn{1}{|c|}{} &
  \multicolumn{1}{l|}{} &
  \multicolumn{1}{c|}{\begin{tabular}[c]{@{}c@{}}LSTM\\ -M1\end{tabular}} &
  \multicolumn{1}{c|}{\begin{tabular}[c]{@{}c@{}}LSTM\\ -M2\end{tabular}} &
  \multicolumn{1}{c|}{\begin{tabular}[c]{@{}c@{}}LSTM\\ -M3\end{tabular}} &
  \multicolumn{1}{c|}{\begin{tabular}[c]{@{}c@{}}LSTM\\ -M4\end{tabular}} &
  \multicolumn{1}{c|}{\begin{tabular}[c]{@{}c@{}}BERT\\ -M1\end{tabular}} &
  \multicolumn{1}{c|}{\begin{tabular}[c]{@{}c@{}}BERT\\ -M2\end{tabular}} &
  \multicolumn{1}{c|}{\begin{tabular}[c]{@{}c@{}}BERT\\ -M3\end{tabular}} &
  \begin{tabular}[c]{@{}c@{}}BERT\\ -M4\end{tabular} \\ \hline
\textit{Dw\_loc1} &
  83 &
  \multicolumn{1}{c|}{2} &
  \multicolumn{1}{c|}{-} &
  \multicolumn{1}{c|}{1} &
  \multicolumn{1}{c|}{38} &
  \multicolumn{1}{c|}{-} &
  \multicolumn{1}{c|}{1} &
  \multicolumn{1}{c|}{5} &
  12 \\ \hline
\textit{Dw\_loc2} &
  37 &
  \multicolumn{1}{c|}{-} &
  \multicolumn{1}{c|}{-} &
  \multicolumn{1}{c|}{-} &
  \multicolumn{1}{c|}{5} &
  \multicolumn{1}{c|}{-} &
  \multicolumn{1}{c|}{-} &
  \multicolumn{1}{c|}{-} &
  3 \\ \hline
\textit{Hr\_btm} &
  16 &
  \multicolumn{1}{c|}{-} &
  \multicolumn{1}{c|}{-} &
  \multicolumn{1}{c|}{1} &
  \multicolumn{1}{c|}{-} &
  \multicolumn{1}{c|}{-} &
  \multicolumn{1}{c|}{-} &
  \multicolumn{1}{c|}{-} &
  11 \\ \hline
\textit{Hr\_top} &
  26 &
  \multicolumn{1}{c|}{-} &
  \multicolumn{1}{c|}{-} &
  \multicolumn{1}{c|}{-} &
  \multicolumn{1}{c|}{-} &
  \multicolumn{1}{c|}{-} &
  \multicolumn{1}{c|}{-} &
  \multicolumn{1}{c|}{-} &
  - \\ \hline
\textit{ID\_stud} &
  56 &
  \multicolumn{1}{c|}{-} &
  \multicolumn{1}{c|}{-} &
  \multicolumn{1}{c|}{-} &
  \multicolumn{1}{c|}{3} &
  \multicolumn{1}{c|}{-} &
  \multicolumn{1}{c|}{-} &
  \multicolumn{1}{c|}{-} &
  - \\ \hline
\textit{ID\_wall} &
  45 &
  \multicolumn{1}{c|}{-} &
  \multicolumn{1}{c|}{} &
  \multicolumn{1}{c|}{-} &
  \multicolumn{1}{c|}{3} &
  \multicolumn{1}{c|}{-} &
  \multicolumn{1}{c|}{-} &
  \multicolumn{1}{c|}{-} &
  - \\ \hline
\textit{O} &
  3,021 &
  \multicolumn{1}{c|}{-} &
  \multicolumn{1}{c|}{-} &
  \multicolumn{1}{c|}{1} &
  \multicolumn{1}{c|}{3} &
  \multicolumn{1}{c|}{-} &
  \multicolumn{1}{c|}{-} &
  \multicolumn{1}{c|}{1} &
  - \\ \hline
\textit{St\_loc1} &
  259 &
  \multicolumn{1}{c|}{-} &
  \multicolumn{1}{c|}{-} &
  \multicolumn{1}{c|}{3} &
  \multicolumn{1}{c|}{39} &
  \multicolumn{1}{c|}{-} &
  \multicolumn{1}{c|}{-} &
  \multicolumn{1}{c|}{-} &
  4 \\ \hline
\textit{St\_loc2} &
  94 &
  \multicolumn{1}{c|}{-} &
  \multicolumn{1}{c|}{-} &
  \multicolumn{1}{c|}{-} &
  \multicolumn{1}{c|}{2} &
  \multicolumn{1}{c|}{-} &
  \multicolumn{1}{c|}{-} &
  \multicolumn{1}{c|}{-} &
  2 \\ \hline
\textit{Vr\_md} &
  171 &
  \multicolumn{1}{c|}{-} &
  \multicolumn{1}{c|}{2} &
  \multicolumn{1}{c|}{4} &
  \multicolumn{1}{c|}{16} &
  \multicolumn{1}{c|}{-} &
  \multicolumn{1}{c|}{-} &
  \multicolumn{1}{c|}{-} &
  - \\ \hline
\textit{dim} &
  19 &
  \multicolumn{1}{c|}{-} &
  \multicolumn{1}{c|}{-} &
  \multicolumn{1}{c|}{-} &
  \multicolumn{1}{c|}{4} &
  \multicolumn{1}{c|}{-} &
  \multicolumn{1}{c|}{-} &
  \multicolumn{1}{c|}{-} &
  1 \\ \hline
\textit{length} &
  34 &
  \multicolumn{1}{c|}{-} &
  \multicolumn{1}{c|}{-} &
  \multicolumn{1}{c|}{1} &
  \multicolumn{1}{c|}{-} &
  \multicolumn{1}{c|}{-} &
  \multicolumn{1}{c|}{-} &
  \multicolumn{1}{c|}{-} &
  1 \\ \hline
\textit{width} &
  34 &
  \multicolumn{1}{c|}{-} &
  \multicolumn{1}{c|}{-} &
  \multicolumn{1}{c|}{-} &
  \multicolumn{1}{c|}{31} &
  \multicolumn{1}{c|}{-} &
  \multicolumn{1}{c|}{-} &
  \multicolumn{1}{c|}{-} &
  9 \\ \hline
TOTAL &
  3,895 &
  \multicolumn{1}{c|}{2} &
  \multicolumn{1}{c|}{2} &
  \multicolumn{1}{c|}{11} &
  \multicolumn{1}{c|}{144} &
  \multicolumn{1}{c|}{-} &
  \multicolumn{1}{c|}{1} &
  \multicolumn{1}{c|}{6} &
  43 \\ \hline
\end{tabular}
\end{table*}

\begin{figure}[h!]
    \centering
    \includegraphics[width=.5\textwidth]{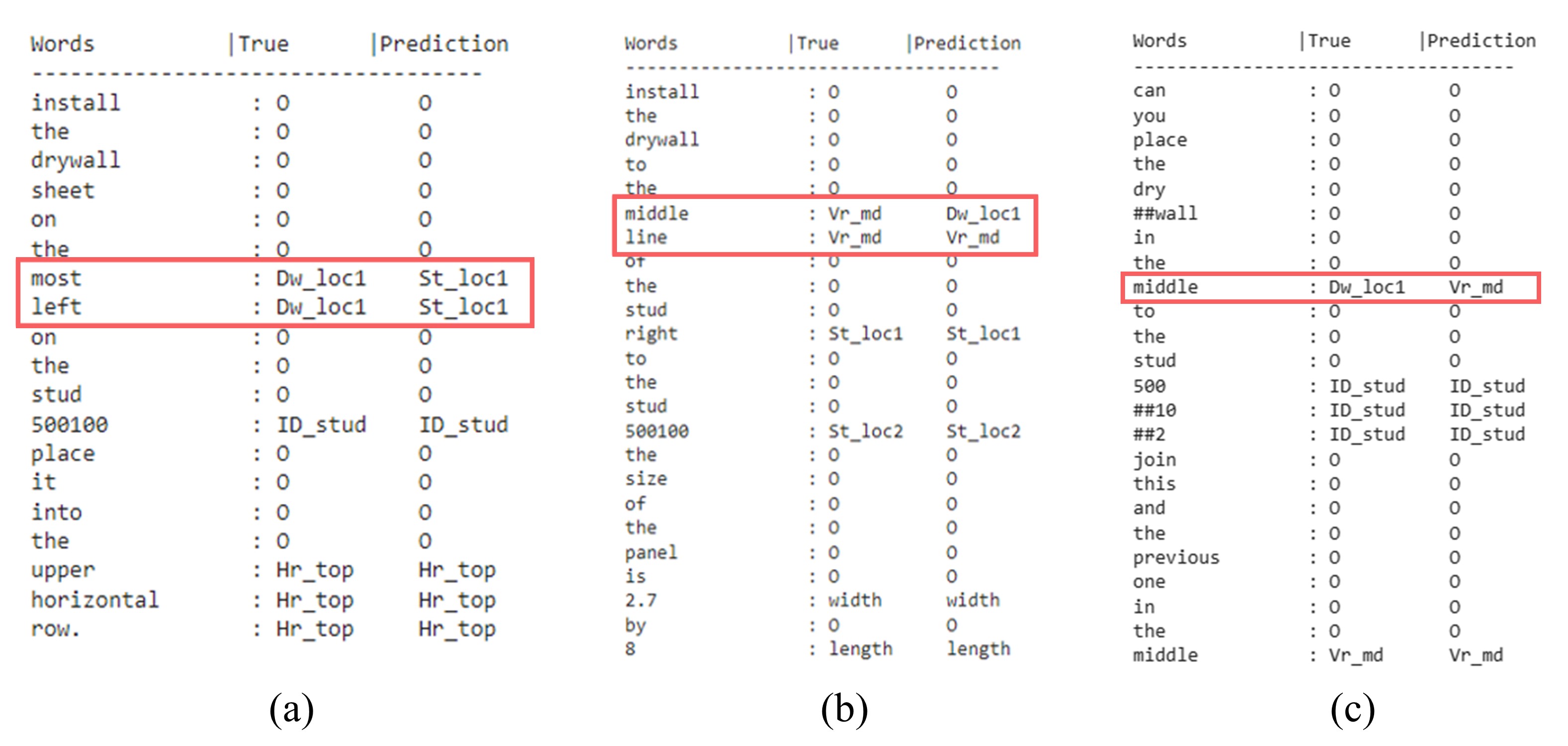}
    \caption{Examples of errors in: (a) LSTM-M1, (b) LSTM-M2, and (c) BERT-M2.}
   \end{figure}
LSTM-M4 and BERT-M4, which were trained with a limited amount of data, had 144 and 43 incorrect predictions, respectively. Most incorrect predictions occurred in the \emph{Dw\_loc1} category. LSTM-M4 displayed a high number of prediction errors for the \emph{Dw\_loc1}, \emph{St\_loc1}, \emph{Vr\_md}, and width labels. In contrast, BERT-M4 had far fewer prediction errors in these categories, which is attributed to its token-level classification approach and pre-trained BERT original version. However, unlike other models, BERT-M4 exhibited a high error rate in predicting \emph{Hr\_btm}, with all corresponding words being incorrectly predicted as \emph{Hr\_top}. This suggests that when BERT models are trained with small datasets, placement methods may be mispredicted, leading to incorrect positioning of the target panel on the stud by the robot. In the test dataset, BERT-M1, which exhibited the best performance, achieved a word-level accuracy of 99.95\% with two incorrect predictions and an instruction-level accuracy of 99.37\% with one error. The error occurred when the values corresponding to width and length were incorrectly predicted as length and width, respectively.

In the test using the BERT-M1 on the Google Colab platform, which offers the use of free GPU, the results showed that the average prediction time for one instruction was about 0.025 seconds. The 158 test data can be categorized into four groups based on the number of sentences: 46 one-sentence instructions, 74 two-sentences instructions, 27 three-sentences instructions, and 11 four-sentences instructions. The average prediction time of each group was 0.0224 seconds, 0.0176 seconds, 0.0324 seconds, and 0.0606 seconds, respectively. As the number of sentences in a single instruction increased, the analysis time tended to increase as well. In other words, time performance is better when the number of sentences is smaller. However, the absolute value was negligible across all sentence groups, showing the effectiveness of the NLU module.

Using studs and drywall panels introduced in the case study, drywall panels can be placed in three different types as shown in Fig. 15. The layouts in Fig. 15(a) and Fig. 15(b) use one unique panel A and one unique panel B, and two standard panels installed vertically and horizontally, respectively. In the layout in Fig. 15(c), two types of distinct panels are placed vertically. Drywall installation is demonstrated based on the outputs of the NLU module and the IM module for three drywall layouts. The input data of the NLU module were selected from the test dataset.
\begin{figure}[h!]
    \centering
    \includegraphics[width=.4\textwidth]{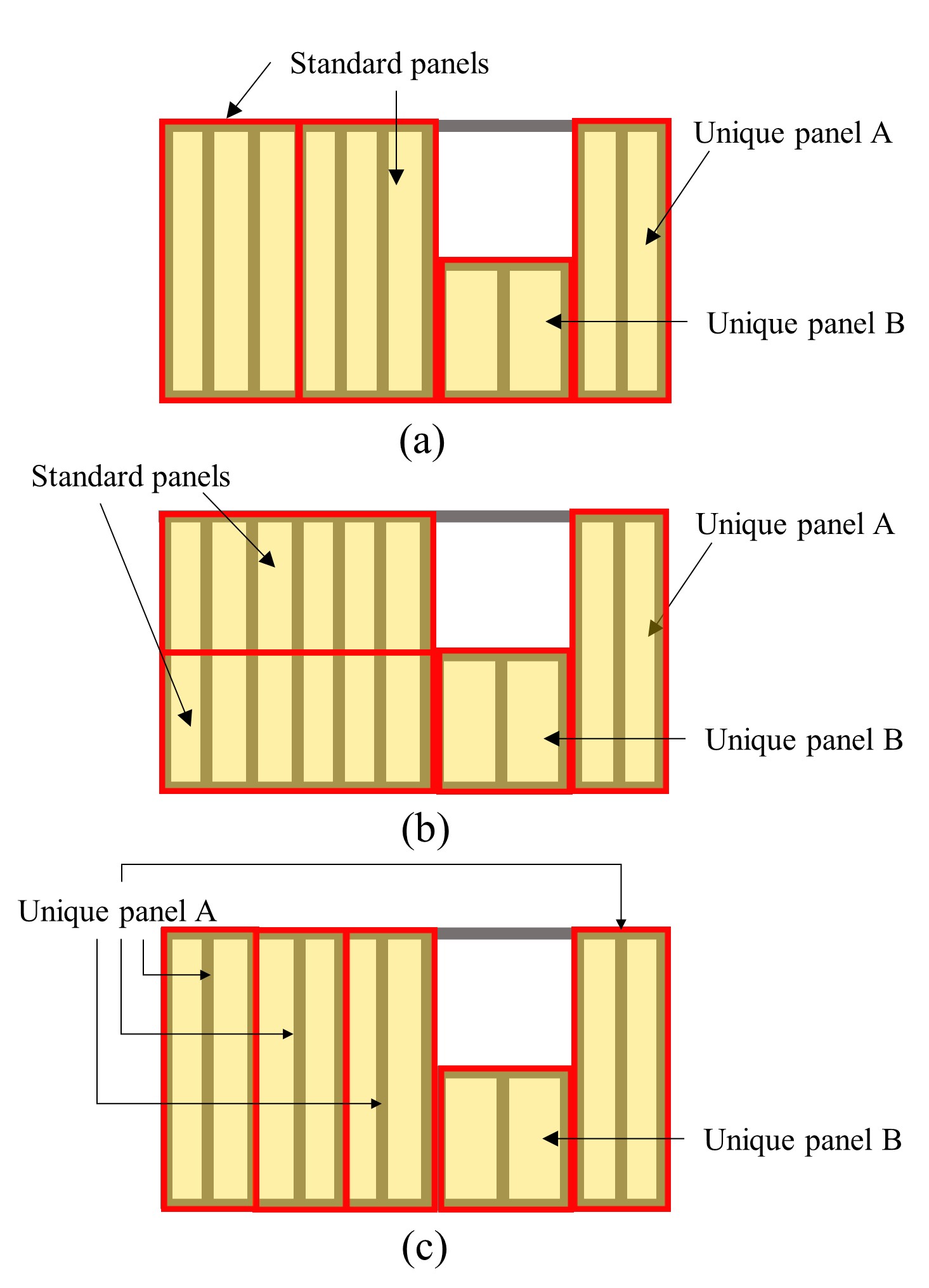}
    \caption{Three drywall layouts: (a) layout 1; (b) layout 2; (c) layout 3.}
   \end{figure}
Demonstration results for the layout 1 are shown in the Fig. 16. Figs. 16(a)-(d) show a pair of a natural language instruction and how the KUKA robot successfully placed a panel for each instruction. As a result of IM for the instruction in Fig. 16(a), the drywall panel 500320 and the stud 500100 were determined as the target and the final location, respectively. The target panel was installed perpendicular to the left line of the stud. The first row of the action history table in Fig. 16(c) shows this result. 
\begin{figure*}[h!]
    \centering
    \includegraphics[width=1.0\textwidth]{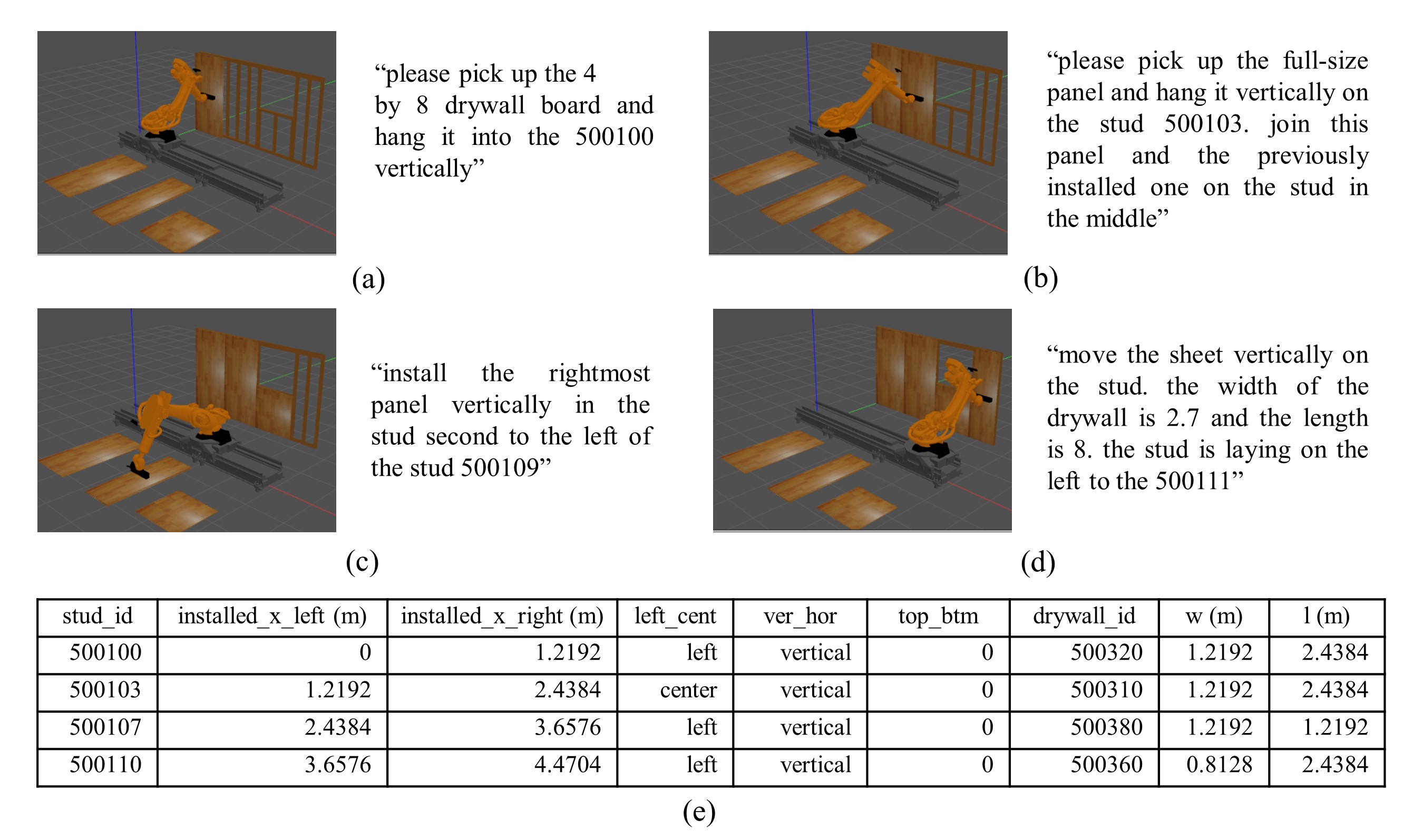}
    \caption{Examples of drywall installation for the layout 1: (a)-(d) show a robot installing drywall panels based on natural language instructions; (e) is the action history table.}
   \end{figure*}
As shown in Fig. 16(b), the drywall panel was installed vertically on the center line of the stud because \emph{Vr\_md} was predicted as a result of the NLU module for the second sentence of the language instruction. The second row of the fourth and fifth columns in Fig. 16(e) shows this result. In Fig. 16(c) and Fig. 16(d), “second to the left” and “left” were tagged as \emph{St\_loc1}, and “500109” and “500111” were tagged as \emph{St\_loc2} in the NLU module. The rules of the IM module shown in Fig. 11 determined the stud 500107 and the stud 500110 as the final location for the third and fourth instructions, respectively. According to the action history table about the output of the IM, the robot installed drywall panels onto the stud walls.

Fig. 17 and Fig. 18 show the natural language instructions and demonstration results for layout 2 and layout 3. As shown in both figures, the robot successfully installed drywall panels by extracting correct information for pick-and-place operations from the NLU and IM modules.

\begin{figure}[h!]
    \centering
    \includegraphics[width=.5\textwidth]{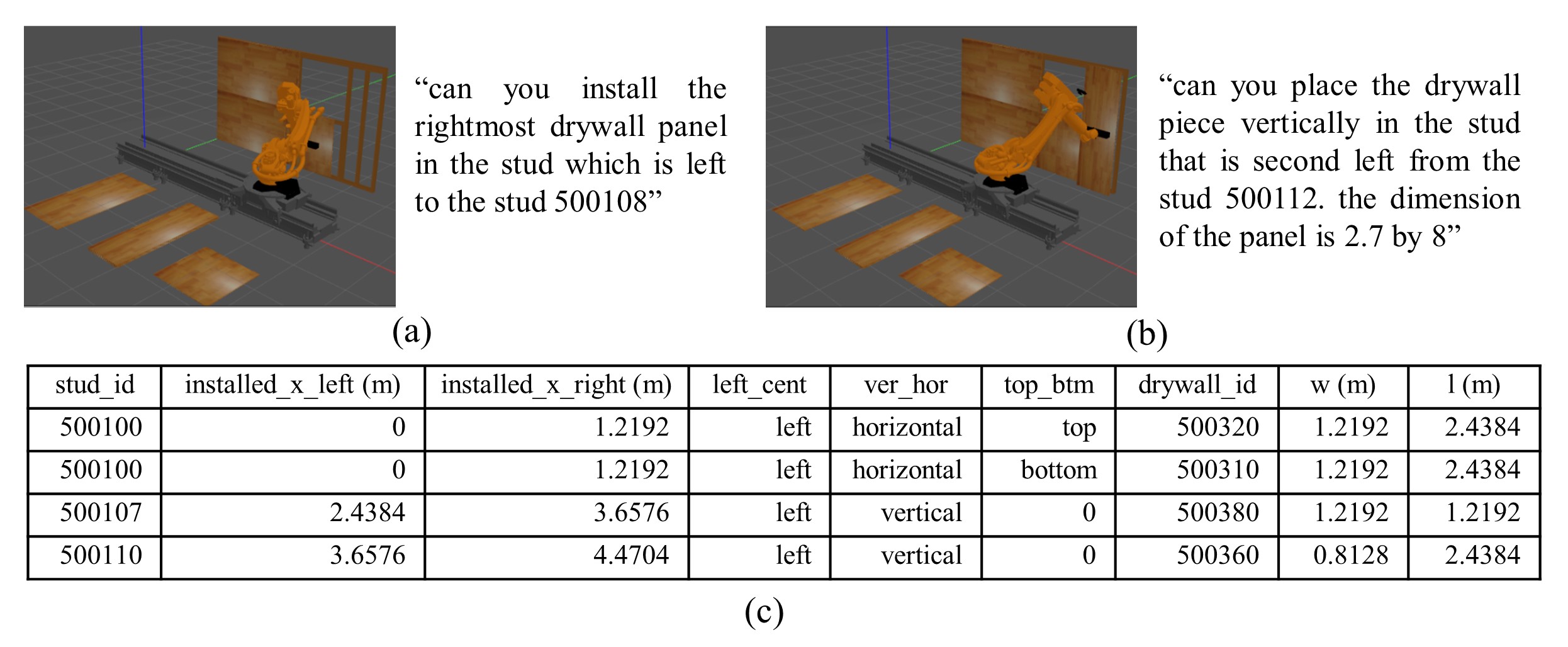}
    \caption{Examples of drywall installation for the layout 2: (a) and (b) are corresponding to the third and fourth placement, respectively; (c) is the recorded action history.}
   \end{figure}

\begin{figure}[h!]
    \centering
    \includegraphics[width=.5\textwidth]{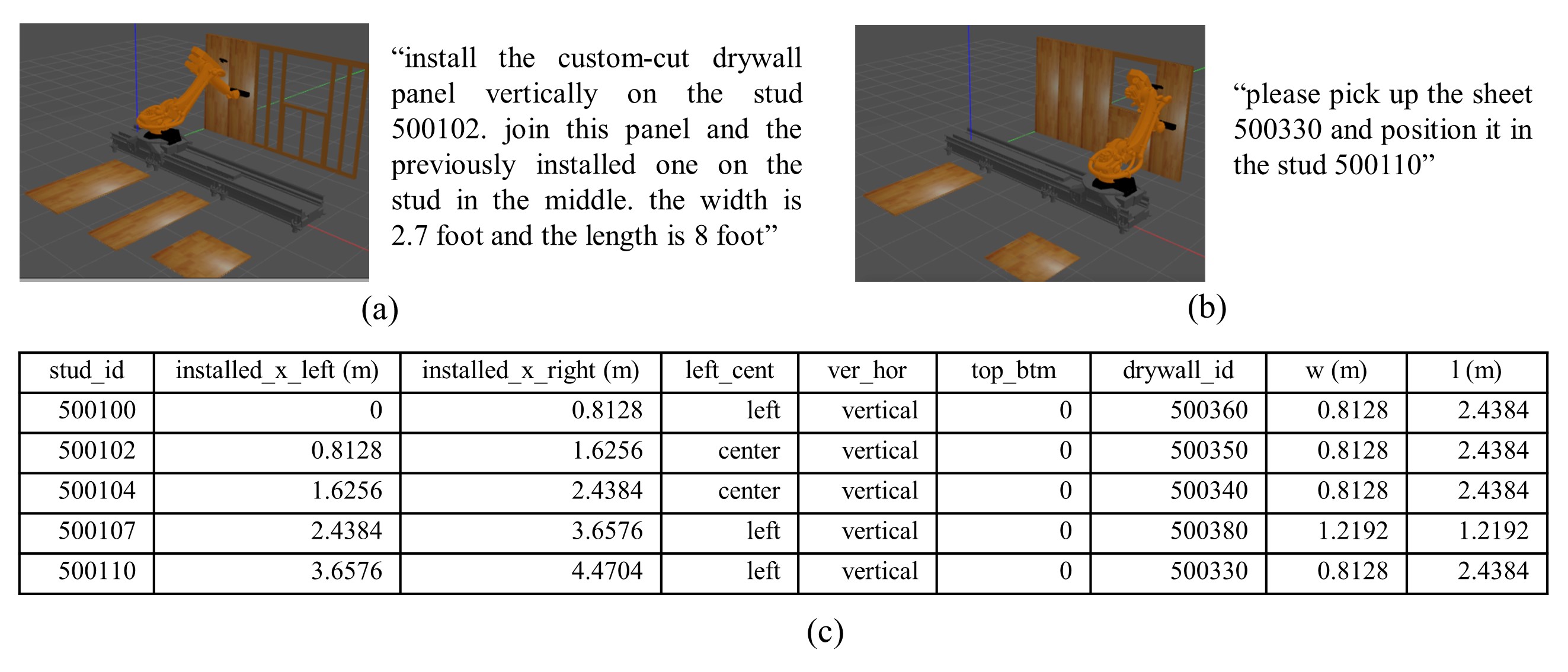}
    \caption{Examples of drywall installation for the layout 3: (a) and (b) are corresponding to the second and fifth placement, respectively; (c) is the recorded action history.}
   \end{figure}
   
\subsection{Co-reference issue}
This study focused on words distinctly characterizing targets and destinations when establishing annotation rules, rather than all words denoting the targets and destinations. This annotation strategy was chosen due to the insufficiency of generic words like drywall, stud or pronouns in clearly distinguishing among multiple panels or studs. However, co-reference issues are crucial for robots to thoroughly interpret human instructions. Thus, additional experiments addressing co-reference issues were conducted using BERT to evaluate the impacts of the co-reference issues in this study.

The dataset was re-annotated with two additional labels: \emph{Trg} and \emph{Dst}, representing a target and destination, respectively. For instance, in a three-sentences instruction “Please move the wall panel and move it on the stud 500100. Place it to the upper horizontal row. The dimension of the drywall is 4 by 8”, ‘wall panel’ in the first sentence, ‘it’ in the second sentence, and ‘drywall’ in the third sentence were annotated as \emph{Trg} while ‘stud’ in the first sentence was annotated as \emph{Dst}. BERT was trained following the same procedure as the prior experiments with variations in the volume of training data. Fig. 19 presents the training accuracy for the re-annotated datasets comprising 316, 632, 948, and 1,268 instructions.
\begin{figure}[h!]
    \centering
    \includegraphics[width=.5\textwidth]{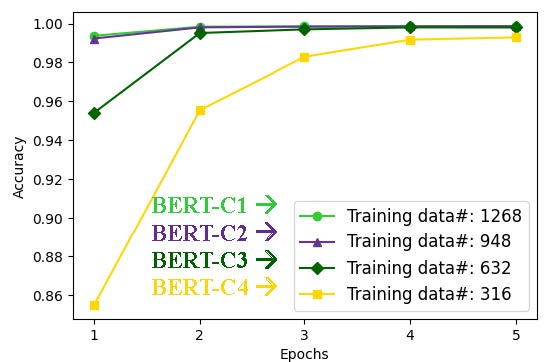}
    \caption{Training accuracy on the re-annotated dataset.}
   \end{figure}
   
The insights from Fig. 13(b) and Fig. 19 reveal that the impact of the co-reference issue on training accuracy is not significant in this study. Initially, in epoch 1, the BERT-C models exhibited lower accuracy in comparison to the BERT-M models. However, as training progressed up to epoch 5, the training accuracy of both BERT-C and BERT-M models converged and became similar. Table 5 presents a comprehensive summary of the performance of the trained models on the validation dataset. It can be observed that BERT-C models, which considered co-reference issues, displayed slightly lower performance compared to the BERT-M models, which did not consider co-reference. However, with a large amount of training data, both BERT-C1 and BERT-C2 achieved accuracy close to 100\%. These findings indicate that while co-reference issues may have a minor impact on performance, the BERT models trained with co-reference consideration can still achieve high accuracy when provided with a large amount of training data.
\begin{table}[]
\caption{Model performance on validation dataset with co-reference issues.}
\centering
\begin{tabular}{|l|cc|cc|}
\hline
\multicolumn{1}{|c|}{\multirow{2}{*}{Model}} & \multicolumn{2}{c|}{Result 1} & \multicolumn{2}{c|}{Result 2} \\ \cline{2-5} 
\multicolumn{1}{|c|}{} & \multicolumn{1}{c|}{$N_{w}$} & \emph{Acc\_word}     & \multicolumn{1}{c|}{$N_{l}$}  & \emph{Acc\_inst}       \\ \hline
BERT-C1                & \multicolumn{1}{c|}{2}    & 99.95\% & \multicolumn{1}{c|}{2}  & 98.73\% \\ \hline
BERT-C2                & \multicolumn{1}{c|}{2}    & 99.95\% & \multicolumn{1}{c|}{2}  & 98.73\% \\ \hline
BERT-C3                & \multicolumn{1}{c|}{14}   & 99.64\% & \multicolumn{1}{c|}{11} & 93.04\% \\ \hline
BERT-C4                & \multicolumn{1}{c|}{62}   & 98.41\% & \multicolumn{1}{c|}{44} & 72.15\% \\ \hline
\end{tabular}
{\small
        \begin{tablenotes}
        \item[a] $N_{w}=$ the number of incorrect prediction of words.
        \item[b] \emph{Acc\_word} $=(3895-N_{w})/3895$
        \item[c] $N_{l}=$ the number of language instructions including incorrect prediction
        \item[d] \emph{Acc\_inst} $=(158-N_{l})/158$
        \end{tablenotes}
        }
\end{table}

\section{Discussion}
This paper presented a framework of a natural language-enabled HRC system that consists of three steps: natural language understanding, information mapping, and robot control. The proposed approach enables human workers to interact with construction robots using natural language instructions and building component information. The proposed system was validated through a case study on drywall installation and BERT-M1 achieved a highest accuracy of 99.37\% at instruction-level for the 158 test data in the NLU module. Even with a small amount of training data, BERT achieved an instruction-level accuracy close to 80\%, suggesting that it is an effective approach for analyzing natural language instructions in the context of construction robotics. However, it should be noted that BERT-based models may require more training time compared to BiLSTM-based models \cite{ezen2020comparison}. Therefore, if the amount of available data is sufficient, it may be worthwhile to consider using the BiLSTM-CRF model, which has shown similar performance to BERT for tagging tasks in this study. In the IM and RC module, it is observed that drywall installation tasks were performed successfully through natural interaction using language instructions. This study clearly demonstrates that the proposed system has significant potential for field implementation to achieve natural interaction with robots in construction.

Even though the proposed method achieved high performance on the given datasets, there are still some challenges that must be addressed. The proposed method used text data as input and a virtual robot digital twin in the experiments instead of using voice data with a real robot deployed on an actual worksite. In the real world, the background noise on-site can interfere in the recognition of the spoken language instructions, which can result in low accuracy in the sequence labeling tasks. In addition, Hatori et al. \cite{hatori} found that the grasping ability of a robot introduced some problems even though the detection of a destination box and a target object achieved high accuracy. In a future study, the authors will explore how spoken language instructions and a physical robot affect the result of the communication between human workers and robots on construction sites. 

Second, the proposed framework relies entirely on the output of the NLU module to generate the final command in the IM module. Consequently, if the NLU module's prediction is incorrect, the IM module's output will be incorrect as well. Future studies can explore integrating the NLU and IM modules and utilizing natural language instructions and building component information as inputs for training together. This could potentially improve the framework's overall accuracy and robustness. Third, the case study was conducted in a single stud structure. In the environment setting, a fixed perspective was used to describe locations of the panels and studs. In future work, the proposed approach can be improved by updating the proposed system for complex structures and changing the perspective of human workers. 

Finally, bidirectional communication was not considered in the proposed system. It implies that human workers are unable to intervene in robot tasks or provide new plans when the robot encounters difficulties for higher level of HRC. This limitation highlights the need for more complicated communication protocols that require a deeper understanding of human-robot interaction. To address this, the authors will consider bidirectional communication in a future study to improve the proposed system and increase the level of natural interaction with construction robots.

\section{Conclusion}
This study made several contributions: the research laid the foundation for natural interaction with construction robots by using natural language instructions. To our best knowledge, it is the first study to demonstrate interaction with construction robots using natural language instructions and building component information. A demonstration of the proposed system using natural language instructions showed the potential of HRC through speech channels in construction. We extracted information about target objects, destinations, and placement orientation that can be applied to other pick-and-place operations in construction tasks, such as ceiling tile installation, wall tile installation, or bricklaying. Even though, the application of the framework we proposed was demonstrated through a drywall installation, the framework itself consisting of three modules (NLU, IM, and RC) is generalizable and adaptable to any pick-and-place construction task making this technical contribution broadly applicable.

Second, to address the lack of an existing dataset suitable for drywall installation, a natural language instruction dataset was created based on human interactions and work observed in construction videos and related studies. The dataset stands out due to its fine-grained annotation as it was meticulously annotated to deal with the necessary information for pick-and-place operations including unique characteristics such as IDs, dimensions, or locations. This annotation process enhanced the quality and depth of the labeled data, making our dataset a valuable resource for advancing research in the field of construction-related natural language processing.

Third, the proposed system facilitates interaction with the robot by using the information available in the construction projects. By mapping building component information and analyzed language instructions, human operators can give language instructions to a robot in a shorter or more intuitive way. We believe that this approach significantly contributes to the development of a practical and efficient human-robot collaboration system on construction sites.

Finally, two different language models, which are BiLSTM-CRF and BERT, were trained by labels reflecting characteristics of construction activities. The results from two different language models were compared and the resulting insights were discussed. It was found that both existing language models worked well with the newly generated dataset. In addition, BERT was an effective approach even when there is limited training data available. Even when trained with 632 instructions, it achieved an instruction-level accuracy of 96\% in validation set. This has important implications for the construction industry, where there is a lack of data for natural language instructions. By leveraging pre-trained models like BERT and fine-tuning them, it is possible to overcome this challenge and achieve high levels of accuracy. In addition, this study showed that BERT achieved high accuracy when trained with a large amount of data while taking co-reference issues into account. Overall, the proposed system demonstrated significant potential in utilizing natural interaction using spoken language instructions in human robot collaboration in construction. It can allow human workers to easily learn how to collaborate with robots through the natural and intuitive interface.

\section{Acknowledgments}
The work presented in this paper was supported financially by two United States National Science Foundation (NSF) Awards: 2025805 and 2128623. The support of the NSF is gratefully acknowledged.

\bibliographystyle{unsrt}
\bibliography{mybib}

\end{document}